\documentclass[lettersize,journal]{IEEEtran}
\usepackage{amsmath,amsfonts}
\usepackage{algorithmic}
\usepackage{algorithm}
\usepackage{array}
\usepackage[caption=false,font=normalsize,labelfont=sf,textfont=sf]{subfig}
\usepackage{textcomp}
\usepackage{stfloats}
\usepackage{url}
\usepackage{verbatim}
\usepackage{graphicx}
\usepackage{cite}
\usepackage{booktabs}
\usepackage{multirow}
\usepackage{tabularx}
\usepackage{array}
\usepackage{enumitem}
\usepackage{caption}
\usepackage{hyperref}
\begin{document}

\title{Out-of-distribution Detection in \\ Medical Image Analysis: A survey}

\author{
    \IEEEauthorblockN{Zesheng Hong\IEEEauthorrefmark{1}, Yubiao Yue\IEEEauthorrefmark{2},Yubin Chen\IEEEauthorrefmark{3}, Lele Cong\IEEEauthorrefmark{11}, Huanjie Lin\IEEEauthorrefmark{4}, Yuanmei Luo\IEEEauthorrefmark{5}, Mini Han Wang\IEEEauthorrefmark{6}, Weidong Wang\IEEEauthorrefmark{7}, Jialong Xu\IEEEauthorrefmark{2}, Xiaoqi Yang\IEEEauthorrefmark{8},Hechang Chen\IEEEauthorrefmark{10}, Zhenzhang Li\IEEEauthorrefmark{9}, Sihong Xie\IEEEauthorrefmark{1}}
    
    \IEEEauthorblockA{\IEEEauthorrefmark{1} Artificial Intelligence Thrust, The Hong Kong University of Science and Technology (Guangzhou), China}
    
    \IEEEauthorblockA{\IEEEauthorrefmark{2} School of Biomedical Engineering, Guangzhou Medical University, China}
    
    \IEEEauthorblockA{\IEEEauthorrefmark{3} Department of Otolaryngology Head and Neck Surgery, the Third Affiliated Hospital of Sun Yat-Sen University, Guangzhou, 510630, P. R.China}

    \IEEEauthorrefmark{11}{\IEEEauthorrefmark{11} Department of Neurology, China-Japan Union Hospital of Jilin University, Changchun, China}
    
    \IEEEauthorblockA{\IEEEauthorrefmark{4} Department of Radiology, The Second Affiliated Hospital of Guangzhou Medical University}
    
    \IEEEauthorblockA{\IEEEauthorrefmark{5} Laboratory Department, The people's Hospital of Qingyuan City}
    
    \IEEEauthorblockA{\IEEEauthorrefmark{6} Department of Ophthalmology and Visual Sciences, Chinese University of Hong Kong, Hong Kong, China}
    
    \IEEEauthorblockA{\IEEEauthorrefmark{7} Department of Orthopedics and Traumatology (Sports Injury and Arthroscopy), Panyu District Hospital of Traditional Chinese Medicine, Guangzhou, Guangdong, China}
    
    \IEEEauthorblockA{\IEEEauthorrefmark{8} Department of Neonatology, Guangzhou Women and Children's Medical Centre, Guangzhou Medical University, Guangzhou, China}

    \IEEEauthorblockA{\IEEEauthorrefmark{10} School of Artificial Intelligence, Jilin University, Changchun, China}
     
    \IEEEauthorblockA{\IEEEauthorrefmark{9} School of Mathematics and Systems Science, Guangdong Polytechnic Normal University}
     
    \IEEEauthorblockA{Corresponding author:\{sihongxie\}@hkust-gz.edu.cn}
}

\maketitle

\begin{abstract}
Computer-aided diagnostics has benefited from the development of deep learning-based computer vision techniques in these years. Traditional supervised deep learning methods assume that the test sample is drawn from the identical distribution as the training data. However, it is possible to encounter out-of-distribution samples in real-world clinical scenarios, which may cause silent failure in deep learning-based medical image analysis tasks. Recently, research has explored various out-of-distribution (OOD) detection situations and techniques to enable a trustworthy medical AI system. In this survey, we systematically review the recent advances in OOD detection in medical image analysis. We first explore several factors that may cause a distributional shift when using a deep-learning-based model in clinic scenarios, with three different types of distributional shift well defined on top of these factors. Then a framework is suggested to categorize and feature existing solutions, while the previous studies are reviewed based on the methodology taxonomy. Our discussion also includes evaluation protocols and metrics, as well as the challenge and a research direction lack of exploration.
\end{abstract}

\begin{IEEEkeywords}
trustworthy AI, medical image analysis, out-of-distribution detection.
\end{IEEEkeywords}

\section{Introduction}
\IEEEPARstart{T}{raditional} supervised machine learning methods are established based on the naive assumption that the test and training samples are drawn from the same distribution, i.e., in-distribution. However, it doesn’t always hold true in the real world, where out-of-distribution samples may be encountered during inference. For deep learning-based medical image analysis tasks such as disease recognition or organ segmentation, models trained in-house may fail on out-of-distribution samples silently, leading to severe outcomes such as misdiagnosis. Therefore, a trustworthy model must be able to say “I don’t know” when encounter an OOD sample and then take the control to the human expert instead of suggesting an error-prone prediction.\\

To this end, out-of-distribution (OOD) detection in medical image analysis has recently developed and drawn attention in the research community. Existing research is conducted across a range of medical fields, while the evaluations encompass various out-of-distribution settings. Besides, several similar tasks are also explored, such as anomaly detection and uncertainty quantification. However, to our best knowledge, there is no systematic framework that clearly describes and groups these cases, and the terminology used in the literature is diverse and sometimes confusing. \\

In this survey, we focus on out-of-distribution (OOD) detection in two widely studied medical image analysis tasks, namely supervised medical image classification and medical image segmentation, since most of the previous techniques have been developed for them. Our contributions include:\\
\begin{itemize}
    \item \textbf{Problem formulation:} We first explore several factors that may lead to a distributional shift in real clinical scenarios, as well as define and interpret three types of distributional shifts based on these factors, which naturally expands the general OOD detection framework \cite{yang2021} to the field of medical image analysis;
    \item \textbf{Solution framework:} A proper solution framework is proposed to organize the related research from two aspects, namely methodology taxonomy and association with base task model;
    \item \textbf{Study review:} We systematically review the existing studies based on the methodology taxonomy, with a focus on technical details and experiment settings;
    \item \textbf{Evaluation protocols and metrics:} The evaluation protocols, metrics, and test samples corresponding to three proposed OOD types used in the previous studies are summarized;
    \item \textbf{Challenges and future directions:} We also discuss a challenge in this area and identify a research direction that deserves more attention in future work.
\end{itemize}

\section{Preliminary}

\begin{figure}[htbp] 
\centering 
\includegraphics[width=0.4\textwidth]{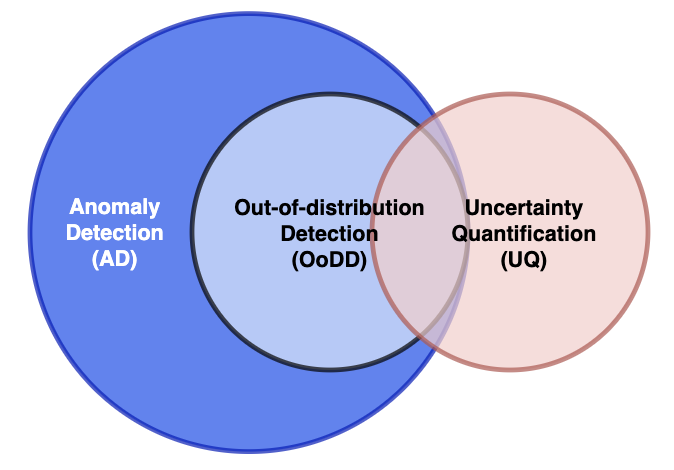} 
\caption{The illustration of relationship between out-of-distribution (OOD) detection, Anomaly Detection (AD), and uncertainty quantification (UQ)} 
\label{fig1} 
\end{figure}

In order to reduce ambiguity and claim the scope of this survey, we first clarify three similar concepts that are easily confused with each other, namely \textit{out-of-distribution (OOD) detection}, \textit{Anomaly Detection (AD)}, and \textit{uncertainty quantification (UQ)}. The relationship between them is illustrated in Fig\ref{fig1}. Besides, it is necessary to clarify the in-distribution before formulating an out-of-distribution (OOD) detection task. To this end, we briefly introduce supervised medical image classification and medical image segmentation in this section. To help readers with no professional background, we also prepare some basic knowledge of several biomedical image types that appear in the involved studies.

\subsection{Out-of-distribution (OOD) detection}\label{subsec: OOD detection}
A supervised deep learning model is trained with a set of instance-label pairs $\{(x_i,y_i)\}_{i=1,...,N}$ to learn match patterns behind them, in the hope that it will generalize well to similar instances. Given a trained model, we term its original learning objective the base task for clarity. Denote the input space  $\mathcal{X}$ and the label (semantic) space $\mathcal{Y}$, an in-distribution is a joint distribution  $P(X,Y)$ over the product space $\mathcal{X} \times \mathcal{Y}$ and is realized by the base task training data pairs. Apart from the base task, a trustworthy model should be capable of identifying the test samples beyond the training in-distribution, namely out-of-distribution (OOD) detection. Due to the distributional mismatch, the OOD sample may exceed the model's perception and thus be an inappropriate input. OOD detection allows marking these problematic inputs and adopting other resorts to handle them properly.

\subsection{Anomaly Detection (AD)}
Anomaly Detection (AD) is a general concept referring to the identification of deviation from normal data \cite{pang2021}. OOD detection can be viewed as a special case of AD when treating the in-distribution samples as normal data. In the research community of deep learning-based medical image analysis, the term \textit{Anomaly Detection} mostly refers to the detection of pathology that deviates from normal healthy images\cite{tschuchnig2022}\cite{fernando2021}\cite{lu2018}\cite{wyatt2022}.
We term this case AD-based pathology detection for clarity. In general, it is achieved through supervised or semi-supervised learning. Supervised learning is to train a classification model with an extremely unbalanced dataset containing both healthy and pathological (abnormal) samples. In contrast, semi-supervised approaches use only healthy samples to train a model capturing the normal pattern and then score the normality during inference to detect pathologies\cite{tschuchnig2022}. AD-based pathology detection is a base task in itself, whereas OOD detection serves as an adjunct to identify unsuitable inputs given a model dedicated to some base task. Thus, we excluded AD-based pathology detection from our survey though some literature \cite{venkatakrishnan2020}\cite{gao2020} call it "out-of-distribution detection" as well.

\subsection{Uncertainty quantification (UQ)}\label{subsec: UQ}

Uncertainty quantification (UQ) is a task that measures predictive uncertainty (PU), i.e., how confident the model feels about the prediction w.r.t. a given input. UQ methods are developed to identify the uncertain sample that requires human review, as well as enable the discovery of the model's deficiency\cite{lambert2024}. Although these techniques are widely utilized to detect OOD samples in medical image analysis \cite{zhang2021} \cite{berger2021} \cite{araujo2023} \cite{thagaard2020} \cite{linmans2020a} \cite{linmans2023b} \cite{mehrtash2020}, we argue UQ and OOD detection are not equivalent and interchangeable concepts. \\

Conventionally, predictive uncertainty (PU) can be factorized into two parts: (1) aleatoric uncertainty (AU) and (2) epistemic uncertainty (EU). The aleatoric uncertainty, also known as data uncertainty, is irreducible as it arises from the inherent properties of data, such as class overlap and noises. In contrast, epistemic uncertainty comes from the lack of knowledge in terms of the underlying model or data, which can be reduced by improving the model's structure, using more training data, or adding valid regularization \cite{zou2023}. Thus, the predictive uncertainty (PU) is modeled as the sum of AU and EU \cite{abdar2021}: $$PU=AU+EU$$
However, another perspective \cite{malinin2018} \cite{nandy2020} accounted for predictive uncertainty (PU) into three divisions: (1) aleatoric uncertainty (AU), (2) model uncertainty (MU), and (3) distributional uncertainty (DU). Here model uncertainty measures the match between the model and training data, while distributional uncertainty arises from the distributional mismatch between the test sample and training set. We argue both model uncertainty (MU) and distributional uncertainty (DU) belong to epistemic uncertainty (EU) as they can be reduced by feeding more data representative of in-distribution samples or OOD samples. Therefore, the predictive uncertainty can be rewritten as: $$PU=AU+\underbrace{MU+DU}_{EU} $$ In other words, high uncertainty may occur in either an in-distribution sample that is hard to predict due to its intrinsic nature (e.g., class overlap) or model deficiency, or an OOD sample. However, most learning-based deterministic UQ methods such as the confidence branch \cite{devries2018}, DUQ \cite{van2020}, and Evidential Deep Learning \cite{sensoy2018} only learn to quantify the uncertainty from in-distribution data, without explicitly considering distributional uncertainty. \cite{ulmer2021} argued that some prevalent UQ methods fail to detect OOD, including Maximum Softmax Probability \cite{hendrycks2016}, MC Dropout \cite{gal2016}, and Deep Ensembles \cite{lakshminarayanan2017}. Besides, recent research \cite{tomani2021} \cite{ovadia2019} \cite{minderer2021} suggested that UQ methods have performance degeneration when distributional shift happens \cite{lambert2024}. \\

Another difference between UQ and OOD detection lies in their evaluation protocols. In fact, the evaluation of UQ is not straightforward simply due to no ground truth of “uncertainty”. Alternatively, it is often achieved by the evaluation of downstream tasks, including OOD detection \cite{lambert2024}. However, UQ can also be evaluated by tasks without involving OOD samples, such as calibration, error detection, segmentation quality control, etc \cite{lambert2024}. Thus, any research dedicated to UQ without considering OOD detection is beyond our scope. For those, please refer to \cite{lambert2024}. 

\subsection{Supervised medical image segmentation}
Supervised medical image classification has been widely applied to a range of computer-aided diagnostic tasks such as distinguishing between malignant and benign lesions, identification of specific pathology, or rating the disease risks \cite{chen2022}. Let $\mathcal{X}$ denote the input space and $\mathcal{Y}$  the label (semantic) space, a medical image $x$ together with its semantic label $y$ lies on the product space $\mathcal{X} \times \mathcal{Y}$. A supervised medical image classification algorithm aims to learn a map ${y=f(x)}$ from training set $D=\{(x_i,y_i)\in \mathcal{X} \times \mathcal{Y}\}_{i=1...,N}$, where $y_i$ is a binary indicator or one-hot encoder representative of a pre-defined class. In this case, the in-distribution is the joint distribution $P(X,Y)$  over $\mathcal{X} \times \mathcal{Y}$ characterized by image-label pairs in the training set.

\subsection{Medical image segmentation}
Medical image segmentation refers to the process of identifying and delineating regions of interest such as lesions, organs, and other substructures. It is achieved by determining the set of pixels (voxels) that belong to these regions rather than the background \cite{chen2022}, thereby can be viewed as pixel (voxel)-level classification. Let $\mathcal{X}$ denote the pixel (voxel) space and $\mathcal{Y}$ the label (semantic) space, any pixel (voxel) $x$ together with its semantic label $y$ lies on the product space $\mathcal{X} \times \mathcal{Y}$. Based on a training set $D=\{(X_i,Y_i)\}_{n=1...,N}$ where $ Y_i $ is a mask indicating label for each pixel within input $ X_i $, a medical image segmentation algorithm aims to learn a map ${y=f(x)}$ for each pixel (voxel), pixels (voxels) with same predicted label form a mask for that category. In this case, the in-distribution is the joint distribution $P(X,Y)$  over $\mathcal{X} \times \mathcal{Y}$  characterized by pixel (voxel)-label pairs in the training set.

\subsection{Biomedical images}
\begin{figure*}[h] 
\centering 
\includegraphics[width=1.0\textwidth]{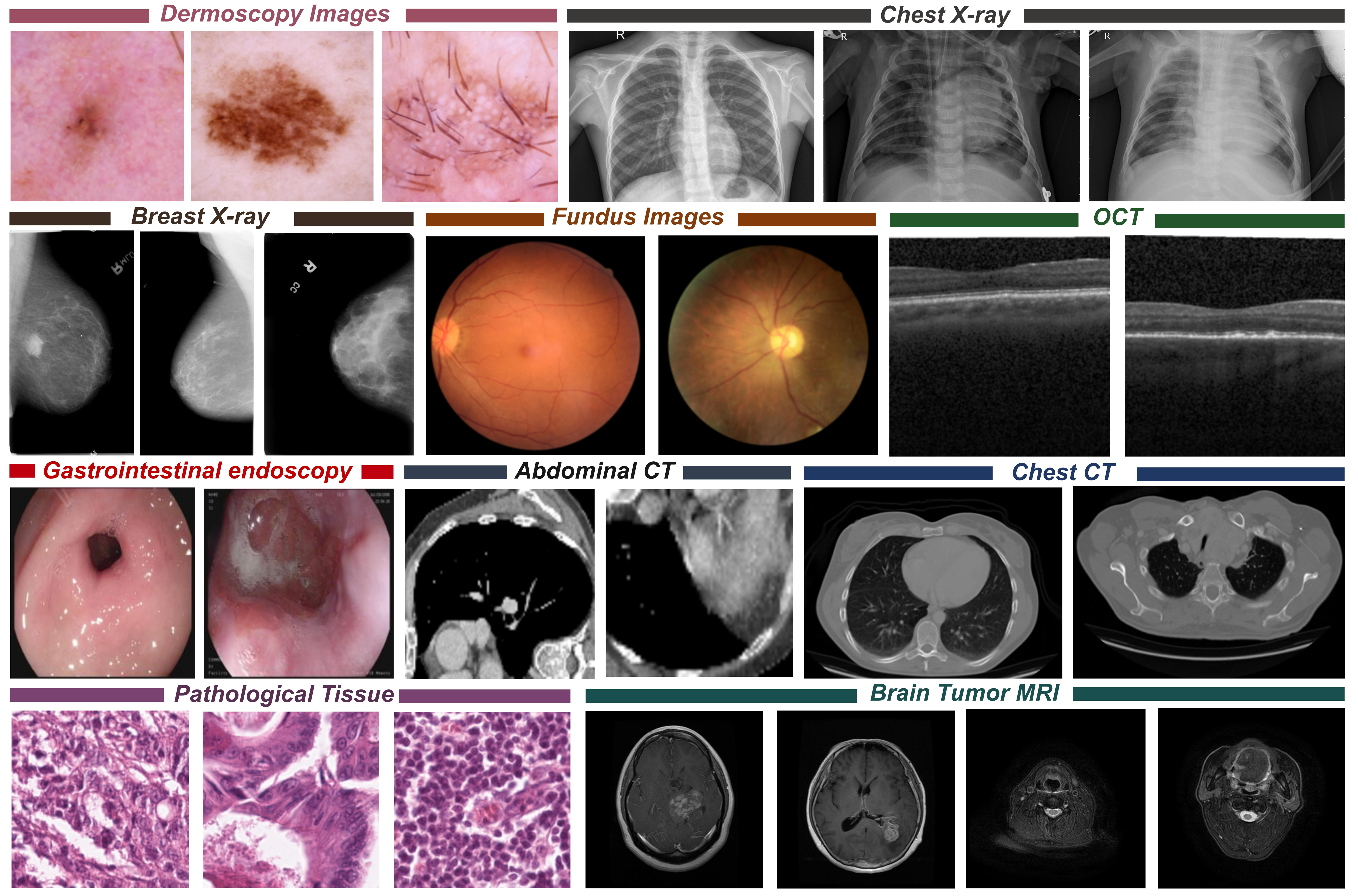} 
\caption{Examples of involved biomedical images} 
\label{fig2} 
\end{figure*}

OOD detection in medical image analysis is studied across a range of medical modalities and image types. We simply introduce some relevant biomedical images to give the reader a primer picture. Please find the example in Fig\ref{fig2}.\\

\noindent\textbf{X-ray images:} X-rays can penetrate through the tissues but would be scattered when encountering bones, which leads to different light exposure in their corresponding imaging area. As the film of X-rays is a negative image (i.e., the darker regions reflect the higher light exposure), the region of bones looks lighter than that of tissues\cite{fernando2021}. Common X-ray images include chest X-ray (CXR) images, Musculoskeletal X-ray images, Mammography images, etc.\\

\noindent\textbf{Fundus images:} A fundus image is a two-dimensional projection of the fundus obtained by a monocular camera, which is appropriate for widespread screening purposes due to the non-invasive acquired manner \cite{li2021}. Fundus images can be used for the diagnosis of common eye diseases, such as glaucoma, cataract, and diabetic retinopathy (DR)\cite{li2021}.\\

\noindent\textbf{Dermoscopy images:} Dermoscopy is a high-resolution skin imaging technique that allows visualization of deeper skin structures by reducing surface reflectance \cite{celebi2019}. The image is captured by skin surface microscopy, which is equipped with high-quality magnifying lens and powerful lighting system. Dermoscopy images are often used to examine pigmented skin lesions, such as Melanoma and Moles.\\

\noindent\textbf{Stained histology slides:} 
A histology slide is a glass slide with the tissue samples fixed upon it, which is typically stained, sectioned, and examined under a microscope\cite{alturkistani2016}. The objective of staining is to color different structures within cells. For example, Hematoxylin, a basic dye employed in this procedure, imparts a bluish color to the nuclei, whereas eosin, another histological stain, imparts a pinkish hue to the cell's nucleus\cite{alturkistani2016}. It is frequently used for diagnosis or classification of cancers.\\

\noindent\textbf{Optical-coherence tomography (OCT):} Optical coherence tomography (OCT) is a technique that allows for the non-contact imaging of the surface and internal microstructure of samples in three dimensions\cite{bouma2022}, which has been popularly used in the diagnosis of retinal diseases such as age-related macular degeneration (AMD)\cite{araujo2023}.\\

\noindent\textbf{Computed tomography (CT):} The Computed tomography (CT) scan is computer-generated cross-sectional images produced through rotating X-rays around a specified body part, which has been proven useful in preventative medicine and cancer screening\cite{patel2021}. It is capable of capturing features in each cross-section and thereby eliminates the superimposition of images in plain films (e.g., X-ray images),  \cite{patel2021}.\\

\noindent\textbf{Magnetic resonance imaging (MRI):} MRI is a non-invasive imaging technique that maps the internal anatomy structures within the body\cite{katti2011}, such as organs, bones, muscles, and blood vessels. Compared with CT scans, the advantage of MRI is it uses radio frequency (RF) radiation instead of electromagnetic radiation, which reduces the exposure-related risk\cite{katti2011}.

\section{Related Work}

\cite{nguyen2015} first discovered the phenomenon that deep neural networks would make an overconfident prediction to out-of-distribution (OOD) data. Since then, out-of-distribution (OOD) detection has been an active field in the research community.\cite{hendrycks2016} pioneered OOD detection, proposing to use Maximum Probability Score (MSP) as a simple baseline. \cite{liang2017} found that simply adding a perturbation to the input in fast gradient sign direction and using temperature scaling improved the OOD detection. \cite{lee2018} suggested modeling the training data features with class conditional Gaussian distribution and testing the OOD sample using Mahalanobis distance. Later, a range of methods and techniques are developed to address this issue, such as Outlier Exposure \cite{hendrycks2018}, energy score \cite{liu2020}, gradient-based GradNorm score \cite{huang2021a}, and generation-based VOS \cite{du2022}.\\

Some aspects other than downstream OOD detection methods also attracted some attention. \cite{fort2021} and \cite{koner2021} focused on the effect of network backbone, demonstrating using a vision transformer (ViT) pre-trained with large-scale datasets can significantly improve the simple MSP \cite{hendrycks2018} and Mahalanobis-based method \cite{lee2018}.\cite{esmaeilpour2022} tried to leverage the multi-modal representation learning, extending the pre-trained language-vision model CLIP to detect OOD. Besides, \cite{huang2021b} argued most works evaluated their methods only on the small, low-resolution datasets, and scaled the OOD detection for datasets with 10-100 times larger label space than previous works.\\

In order to systematically review the recent studies in OOD detection and other related tasks, \cite{yang2021} proposed a well-defined framework named generalized out-of-distribution detection. However, they overlook the studies about OOD detection in medical image analysis. Although \cite{chen2023} reviewed some works in medical image OOD detection, their work is excessively incomprehensive. Besides, it lacks the exploration of problem formulation, a well-organized solution framework, and the discussion of evaluation protocols, challenges, and future directions, as compared to our work.\\

In addition to OOD detection, some surveys are conducted in medical Anomaly Detection (AD) \cite{tschuchnig2022}\cite{fernando2021}. These works only consider AD as a base task while missing the situation where it functions as an auxiliary to improve the reliability of a deep learning-based model, namely OOD detection. \cite{lambert2024} systematically reviewed the uncertainty quantification (UQ) in deep learning-based medical image analysis. Despite some overlaps between UQ and OOD detection, they are not equivalent concepts and should be treated differently, as we argued in \ref{subsec: UQ}. \\

\section{Problem formulation and taxonomy} \label{sec: problem taxonomy}
Distributional shifts can occur across a variety of factors. \cite{ahmed2020} argued that not all distribution shifts should be considered in OOD detection. Instead, a distributional shift should be taken into account only when it occurs on the factors of interest. Let's take object recognition as an example. Usually, the model is trained to classify the identity of the foreground object into several pre-defined semantic classes while not caring about the background. In this case, an image with a pre-defined object located in a novel background should not be an OOD sample. However, the opposite is true in a scene classification task. Motivated by this view, we first analyze the factors on which the distributional shift may happen in clinical scenarios. On top of these factors, we borrow two names from generalized OOD detection framework\cite{yang2021}, \emph{semantic shift}, and \emph{covariate shift}, together with \emph{contextual shift} to characterize three distributional types in medical image analysis. Please note the three terminologies are used hereafter to reduce confusion, though they may have different names, such as “far OOD”, “near OOD”, and “domain shift” in literature. \\

\subsection{Distributional shift factors}\label{subsec:factors}
Referring to \cite{cao2020}  and domain expert's opinion, we summarize seven distributional shift factors of importance in medical image analysis:\\

\noindent\textbf{Modality:} Medical image modality often depends on the acquisition equipment, encompassing but not limited to Magnetic Resonance Imaging (MRI), Computed Tomography (CT),  X-ray, and stained histology slides. Two distinct medical image modalities differ in imaging principles. As a result, the geometric nature and appearance can vary dramatically across two modalities, even for the same object.\\

\noindent\textbf{Area of concern:}  Generally, a medical image classification or segmentation task should be dedicated to a fixed area, such as the brain, chest, skin, or lymph node tissue. Different areas are incomparable due to the differences in their intrinsic anatomical natures.\\

\noindent\textbf{Imaging view:} For some medical images, a shift in imaging views may cause a difference. For example, a posteroanterior (PA) Chest X-ray image is acquired by placing X-rays at the rear of the patient, while an anteroposterior (AP) Chest X-ray image is acquired oppositely. The two exhibit explicable variations due to the patient's positioning and the cone-beam geometry\cite{calli2022}.\\

\noindent\textbf{Image quality:} The existence of image quality issues, such as blurry, poor contrast, and overexposed, may lead to a distributional shift. This arises from incorrect operation or the poor performance of imaging equipment.\\

\noindent\textbf{Acquisition protocols and pre-processing:} The acquisition protocols and pre-processing of medical images tend to vary among medical sites/centers as the principles they follow may be different from each other.\\

\noindent\textbf{Class of target:}  The target in medical images is an analogy to the semantic object in natural image datasets such as CIFAR-10 or ImageNet. Specifically,  it can be a disease, pathology, or cell in medical image classification, as well as a lesion, organ, tissue, or other anatomical structures of interest in medical image segmentation. The input image sometimes contains a novel class of the target, resulting in a distributional shift in semantics.\\

\noindent\textbf{Cohort:}  A cohort refers to a group of individuals or patients who share certain characteristics and are studied together for research or clinical purposes. This group is often selected based on specific criteria, such as age, gender, medical condition, or treatment received. A novel cohort that is unseen in the training set also leads to a distributional shift. For example, a pediatric chest X-ray is OOD to adult chest X-ray image, or a chest X-ray containing artificial implants is OOD to implant-free chest X-ray image.\\

\begin{table*}[h]
\centering
\caption{}
\setlength{\tabcolsep}{3pt}
\renewcommand{\arraystretch}{1.2}
\begin{tabular}{|l|l|l|l|l|l|l|l|}
\hline
 & \multicolumn{1}{c|}{Modality} & \multicolumn{1}{c|}{\begin{tabular}[c]{@{}c@{}}Aera of \\ concern\end{tabular}} & \multicolumn{1}{c|}{\begin{tabular}[c]{@{}c@{}}Imaging \\ view\end{tabular}} & \multicolumn{1}{c|}{\begin{tabular}[c]{@{}c@{}}Image \\ quality\end{tabular}} & \multicolumn{1}{c|}{\begin{tabular}[c]{@{}c@{}}Acquisition protocols \\ and preprocessing\end{tabular}} & \multicolumn{1}{c|}{\begin{tabular}[c]{@{}c@{}}Class \\ of target\end{tabular}} & \multicolumn{1}{c|}{\begin{tabular}[c]{@{}c@{}}Cohort \end{tabular}} \\ \hline
contextual shift & \multicolumn{1}{c|}{\checkmark} & \multicolumn{1}{c|}{\checkmark} &  &  &  &  &  \\ \hline
semantic shift&  &  &  &  &  & \multicolumn{1}{c|}{\checkmark} &  \\ \hline
covariate shift&  &  & \multicolumn{1}{c|}{\checkmark} & \multicolumn{1}{c|}{\checkmark} & \multicolumn{1}{c|}{\checkmark} &  & \multicolumn{1}{c|}{\checkmark} \\ \hline
\end{tabular}
\label{table:1}
\end{table*}

\subsection{OOD detection in medical image analysis}

A generalized OOD detection framework \cite{yang2021} is discussed with respect to a deep learning-based visual recognition model. \cite{yang2021} suggested dichotomizing distributional shift into covariate (sensory) shift and semantic shift. The former refers to the shift that occurs only in the marginal distribution $P(X)$. while the latter means a novel class of object, which leads to the distributional shift in both  $P(Y)$ and  $P(X)$. For example, given a deep learning-based classification model trained with RGB images to distinguish between dog and cat, covariate and semantic shift instances can be a sketch of the dog and an RGB image of the frog, respectively.\\ 

\begin{figure*}[h] 
\centering 
\includegraphics[width=1.0\textwidth]{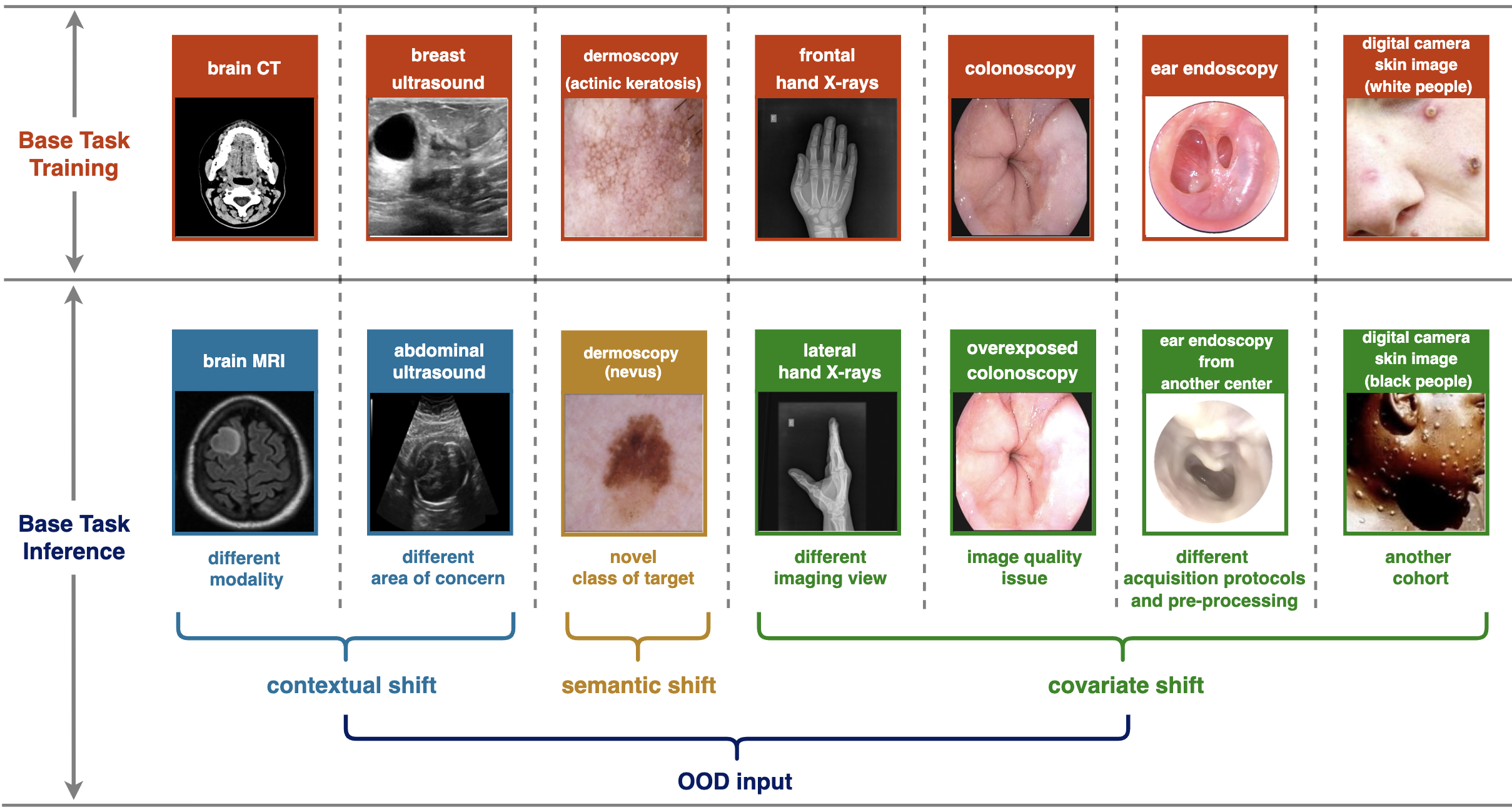}
\caption{Visualization of distributional shift types in medical image analysis} 
\label{fig3} 
\end{figure*}

Despite a perfect fit for natural object recognition, we argue this framework is not appropriate to describe distributional shift cases in medical image analysis. Let's assume the base task is to distinguish between \textit{Lung Opacity} and \textit{Pleural Effusion} based on chest X-ray image, and a set of corresponding chest X-ray image with normal contrast is used to train a deep learning model. Now we consider two different OOD samples: (1) A chest CT slice obtained from a patient with \textit{Lung Opacity}, and (2) A poor-contrast chest X-ray image obtained from a patient with \textit{Lung Opacity}. Based on the definition in \cite{yang2021}, both two can be categorized into covariate shift. However, they differ in the degree of deviation from in-distribution. The former is completely incomparable to in-distribution samples due to the mismatch in modality. In contrast, the latter retains a lot of features representative of in-distribution samples, thereby still having a chance to be predicted correctly. In order to properly adapt the generalized OOD detection framework\cite{yang2021} to clinal scenarios, we propose a taxonomy for OOD detection in medical image analysis and define each category based on the factors discussed before (see Table \ref{table:1}), in the hope that it can well describe all the cases considered in previous studies.\\

\subsubsection{Contextual shift}
The context of a medical image can be roughly described from two aspects, the \textbf{modality} (e.g., X-ray images, CT-scans, histology slides…), and the \textbf{area of concern} (e.g.,  an organ, tissue…). We define contextual shift sample as the input image having inconsistent \textbf{modality} and\textbackslash or \textbf{area of concern} with the training set. Note the non-medical image is a case of contextual shift to any medical images. Usually, a medical image classification or medical image segmentation task only targets a specific context, which means the \textbf{modality} and \textbf{area of concern}  are typically consistent across the training set. However, there is an exception in   \cite{karimi2022} where the model is trained across multiple similar modalities (i.e., CT and MRI) and different organs (e.g., liver, heart...), namely multi-task segmentation. In this case, contextual shift means the input image has a different \textbf{modality} and\textbackslash or \textbf{area of concern} from that of any training samples. The context specifies in which situation the trained model will be correctly used, or in other words, the correct input type. Thus, contextual shift will inevitably lead to a meaningless prediction. For example, it makes no sense to input a chest CT slice into a lung pathology classifier trained with X-ray images, and a tumor segmentation model trained on brain CT must fail to segment COVID-19 lung lesion shown in chest CT. Such a distributional shift often arises from man-made input errors or malicious attacks and should be rejected without any hesitation.\\

\subsubsection{Semantic shift} 
Following the definition in \cite{yang2021}, we define semantic shift as the input image containing a novel \textbf{class of target}. It is common in supervised medical image classification, such as an image with a rare disease that is not defined in the label set. However, we argue semantic shift detection in medical image segmentation is meaningful only when the segmentation object is a lesion instead of organ, tissue, or other anatomical structures whose semantic classes are constant. For example, in \cite{gonzalez2022} the base task is to segment pulmonary Covid-19 lesions in chest CT scans while a set of OOD samples are pulmonary lesions caused by non-Covid pneumonia, bacterial pneumonia, fungal pneumonia, etc. This can be seen as semantic shift in medical image segmentation. A semantic shift arises from incomplete knowledge or a lack of available samples in the training stage. Other than the indication of an erroneous input, the identification of semantic shift samples can benefit the model as well. Once detected, they can be annotated by an oracle (e.g., a physician) and stored in a database. By resorting to continual learning techniques, these samples can be used to update the model's knowledge throughout its entire life span.\\

\begin{table*}[h]
\centering
\caption{}
\setlength{\tabcolsep}{7pt}
\renewcommand{\arraystretch}{1.2}
\begin{tabular}{|c|c|c|c|c|}
\hline
base task & In-distribution Data & contextual Shift & semantic shift & covariate shift \\ \hline
\multirow{11}{*}{\begin{tabular}[c]{@{}c@{}}supervised\\ medical image  classification\end{tabular}} 

 & Chest X-ray (CXR) images & \cite{cao2020}\cite{zhang2021}\cite{graham2023a}\cite{calli2022} & \cite{cao2020}\cite{berger2021} & \cite{cao2020}\cite{calli2022}  \\ \cline{2-5} 
 & Musculoskeletal X-ray images & \cite{zhang2021}\cite{graham2023a} &  &  \\ \cline{2-5}
 & Mammography images &  &  & \cite{tardy2019} \\ \cline{2-5} 
 & Fundus images & \cite{cao2020}\cite{zhang2021}\cite{nandy2021} & \cite{cao2020} & \cite{cao2020}\cite{nandy2021} \\ \cline{2-5} 
 & Dermoscopy \textbackslash digital camera skin images & \cite{pacheco2020}\cite{yasin2020} & \cite{li2022}\cite{roy2022}\cite{pacheco2020}\cite{yasin2020}\cite{kim2022} \cite{combalia2020} & \cite{pacheco2020}\cite{kim2022} \\ \cline{2-5} 
 & Stained Histology Slides & \cite{cao2020}\cite{linmans2023b} & \cite{cao2020}\cite{thagaard2020}\cite{linmans2020a}\cite{linmans2023b} & \cite{linmans2023b} \\ \cline{2-5} 
 & Optical Coherence Tomography (OCT) & \cite{araujo2023} & \cite{araujo2023} &  \\ \cline{2-5}
 & Abdomen CT & \cite{graham2023b} & \multicolumn{1}{l|}{} & \multicolumn{1}{l|}{} \\ \cline{2-5} 
 & Chest CT & \cite{graham2023a} & \multicolumn{1}{l|}{} & \multicolumn{1}{l|}{} \\ \cline{2-5} 
 & Head CT & \cite{graham2023a} & \multicolumn{1}{l|}{} & \multicolumn{1}{l|}{} \\ \cline{2-5} 
 & Breast MRI & \cite{graham2023a} & \multicolumn{1}{l|}{} & \multicolumn{1}{l|}{} \\ \hline 
\multirow{8}{*}{medical image segmentation} 
 & Chest CT & \cite{gonzalez2022} & \cite{gonzalez2022} & \cite{gonzalez2021}\cite{gonzalez2022} \\ \cline{2-5} 
 & Head CT & \cite{graham2022}\cite{graham2023b} &  & \cite{graham2022}\cite{graham2023b} \\ \cline{2-5}
 & Liver T1 MRI &  &  & \cite{woodland2023} \\ \cline{2-5} 
 & Brain cortical plate T2 MRI &  &  & \cite{karimi2022} \\ \cline{2-5} 
 & Prostate T2 MRI &  &  & \cite{mehrtash2020} \\ \cline{2-5}
 & Laparoscopic Cholecystectomy images &  &  & \cite{yang2024} \\ \cline{2-5} 
 & Endoscopic Surgical images &  &  & \cite{yang2024} \\ \cline{2-5} 
 & \begin{tabular}[c]{@{}c@{}}Brain cortical plate-T2 MRI, Prostate-MRI, \\ Heart-MRI, Liver-CT, Liver-MRI\end{tabular} & \cite{karimi2022} &  & {\cite{karimi2022}} \\ \hline
\end{tabular}
\label{table:2}
\end{table*}

\subsubsection{Covariate shift} 
Even without contextual and semantic shifts, an input sample can still deviate far from the training set in covariates \cite{yang2021}. For supervised medical image classification or medical image segmentation, covariates can be explained as \textbf{imaging view}, \textbf{image quality}, \textbf{acquisition protocols and pre-processing}, and \textbf{subject group}, as the shift in these factors would not change the  \textbf{class of target}. We define covariate shift as an input image in which at least one covariate is different from that of any training sample. Covariate shift often results from the inconsistency between data acquisition sources, such as different centers and cohorts. Unlike modality and area of concern, covariates should be diverse across training samples in order to learn a robust model that can generalize well in real clinical cases. Thus, the identification of covariate shift samples can also improve the model's generalization through continuous learning.\\

The three distributional shifts are illustrated in Fig\ref{fig3}). They were considered in previous studies across a wide range of biomedical image types and we summarize the corresponding reference in Table \ref{table:2}. Now OOD detection in medical image analysis can be formulated as follows. Given a medical image classification\textbackslash segmentation model trained with the training set $D$, OOD detection is to find a score function $s(x)$ and a threshold-based detector 
\begin{equation}\label{eq1} 
G(x) = \begin{cases}1,\quad s(x)>\tau\\0,\quad s(x)\leq{\tau}\end{cases}
\end{equation}
so that $G(x) = 1$ if $\hat{y}$ deviates far from the training samples in terms of context, semantics of target, or covariates.

\section{Solution framework}

There are various methods developed to achieve OOD detection in these years. While most of them were initially evaluated in natural image recognition, some have been adapted to the field of medical image analysis. In addition, several studies considered achieving OOD detection in medical image analysis by resorting to UQ techniques. To establish a clear understanding of recent advances in this area, we proposed a solution framework to well organize the existing research from two perspectives, namely \textit{methodology taxonomy} and \textit{association with base task}, shown in Table \ref{table:3}.

\subsection{Methodology taxonomy} \label{sbusec:taxonomy}

First, the methodology of OOD detection in medical image analysis are summarized into five categories based on their principles: \\

\noindent\textbf{Post-hoc feature process:} These methods view the intermediate layer of the base task model as a feature extractor and achieve OOD detection in the latent feature space rather than the output space. Specifically, the representation of each sample is obtained by forwarding the input with the pre-trained model, and some processes are then conducted on these representations to estimate OOD-ness.  The idea is motivated by the fact that a sample far away from the training in-distribution can still obtain a high softmax score\cite{nguyen2015}\cite{hein2019}. These methods were broadly adopted to address OOD detection in medical image analysis due to their ability to obtain OOD scores posterior to the base task training.\\

\noindent\textbf{Learning-free UQ:} As mentioned in the preliminary, the distributional shift is one of the sources of epistemic uncertainty. Thus, uncertainty quantification (UQ) is often treated as the solution to OOD detection in medical image analysis, with low uncertainty indicating a high probability of being OOD. From the view of Bayesian Neural Networks (BNNs), the model parameters are also random variables, which induces a natural factorization of predictive uncertainty as below\cite{malinin2018}: 
$$\mathrm{P}\left(y=\omega_c \mid \boldsymbol{x}^*, \mathcal{D}\right)=\int \underbrace{\mathrm{P}\left(y=\omega_c \mid \boldsymbol{x}^*, \boldsymbol{\theta}\right)}_{\text {AU}} \underbrace{\mathrm{p}(\boldsymbol{\theta} \mid \mathcal{D})}_{\text {EU}} d \boldsymbol{\theta}$$ 
where $\omega_c$, $\boldsymbol{x}^*$, and $\mathcal{D}$ are pre-defined class, input, and the training set, respectively. The aleatoric (data) uncertainty (AU) is described by the posterior distribution over class labels given an input and a fixed set of model parameters, while the epistemic uncertainty (EU) is captured by the posterior distribution over model parameters given a training set\cite{malinin2018}. This framework can be explained as a distribution over all the possible predictive categorical distributions. However, it is intractable to obtain the true model posterior in practice. An alternative is to approximate it through a set of point estimates of predictions generated by MC-dropout  \cite{gal2016} or explicit ensembles \cite{lakshminarayanan2017}\cite{linmans2020a}. Subsequently, the uncertainty is directly quantified by checking the statistics or metrics computed on this set. We term these methods "learning-free" as they do not consider uncertainty estimation as a training objective. Note that the distributional shift is not explicitly taken into account in these methods, which means the OOD detection is achieved by implicitly modeling the distributional uncertainty (DU) through the epistemic uncertainty (EU) \cite{malinin2018}.\\

\noindent\textbf{Learning-based deterministic UQ:} These methods explicitly consider uncertainty modeling during training, which is typically achieved by optimizing a special loss function. Instead of using the statistic or the metric computed on a set of predictions, they output a single deterministic uncertainty in an inference run. However, as there is no available ODD during training, the concept of uncertainty, or confidence, is only learned from in-distribution samples. Thus, we argue that using these methods to detect OOD is equivalent to viewing the hard-to-classify in-distribution sample as a proxy of OOD, with the implicit assumption that the response to the former can generalize to the latter.\\

\noindent\textbf{OOD-aware training:} Introducing a few OOD samples into the training set, these methods attempt to directly learn the discrimination between in-distribution and OOD samples from supervision signals. Furthermore, the model is typically trained in a multi-task fashion, combining the losses of the base task and OOD detection in a certain ratio.
\\

\noindent\textbf{Unsupervised stand-alone detectors:} These methods train a model in an unsupervised manner (i.e., no labels are available during training) using only in-distribution data and anticipate that the model would respond differently to OOD samples. Besides, the model is dedicated to OOD detection and typically stands alone, which means its network architecture, training process, and inference are completely separated from those of the base task. Thus, one can distinguish them from post-hoc feature process based on whether the features retrieved from the base task model are utilized.\\

In section \uppercase\expandafter{\romannumeral 5} and \uppercase\expandafter{\romannumeral 6}, the studies related to OOD detection in supervised medical image classification and medical image segmentation are systematically reviewed based on our taxonomy.

\subsection{Association with base task model}

In addition to the methodology taxonomy, the association between OOD methods and base task model also constitutes a concern in our framework, as it reveals how easily an OOD detection solution can be deployed given a pre-trained model. Specifically, there are three cases in the existing research:\\

\noindent\textbf{Model Reuse:} In this case, there is no additional training process to the base task. Instead, the pre-trained model is only reused to obtain the intermediate feature or final output. Therefore, these methods are able to serve as a plug-and-play tool to equip any pre-trained model.\\

\noindent\textbf{Retraining:} In this case, the pre-trained base task model is retrained from scratch to obtain both the base task prediction and OOD score in a single inference run. Thus, these solutions cannot be directly deployed into a pre-trained model.\\

\noindent\textbf{Independent Training:} These methods require one or multiple training processes independent of the base task. Despite additional computational overhead, one can directly combine them with a pre-trained model to establish a trustworthy system. \\

\noindent Note that the term "training" here only refers to a process involving backpropagation, a simple model fit such as logistic regression or decision tree is not considered a training process.\\

\begin{table*}[h]
\centering
\caption{}
\setlength{\tabcolsep}{1 pt}
\renewcommand{\arraystretch}{1.2}
\begin{tabular}{|c|c|c|c|c|c|c|}
\hline
Base task & Methodology category & Methodology & \begin{tabular}[c]{@{}c@{}}Model reuse\end{tabular} & Retraining & \begin{tabular}[c]{@{}c@{}}Independent training\end{tabular} & Reference \\ \hline
\multirow{23}{*}{\begin{tabular}[c]{@{}c@{}}Supervised medical \\ image classification\end{tabular}} 

& \multirow{7}{*}{\begin{tabular}[c]{@{}c@{}}Post-hoc \\ feature process\end{tabular}} 
 & Simple binary classifier & \checkmark &  &  & \cite{cao2020} \\ \cline{3-7} 
 &  & Mahalanobis & \checkmark &  &  & \cite{cao2020} \cite{berger2021} \cite{araujo2023} \cite{roy2022} \cite{calli2022} \cite{tardy2019} \\ \cline{3-7} 
 &  & Cosine similarity & \checkmark &  &  & \cite{araujo2023} \\ \cline{3-7} 
 &  & Isolation Forests & \checkmark &  &  & \cite{li2022} \\ \cline{3-7} 
 &  & Extreme Value Theorem & \checkmark &  &  & \cite{yasin2020} \\ \cline{3-7} 
 &  & Gram matrix & \checkmark &  &  & \cite{pacheco2020} \\ \cline{3-7} 
 &  & Subset Scaning & \checkmark &  &  & \cite{kim2022} \\ \cline{2-7}
 
 & \multirow{8}{*}{\begin{tabular}[c]{@{}c@{}}Learning-free \\ UQ\end{tabular}} 
 & MSP & \checkmark &  &  & \cite{cao2020} \cite{zhang2021} \cite{berger2021} \cite{araujo2023} \cite{roy2022}\cite{calli2022} \cite{kim2022} \cite{tardy2019} \\ \cline{3-7}
 &  & Entropy & \checkmark &  &  & \cite{linmans2020a} \cite{linmans2023b} \cite{tardy2019} \\ \cline{3-7} 
 &  & Temperature Scaling & \checkmark &  &  & \cite{berger2021} \cite{araujo2023} \cite{linmans2023b} \\ \cline{3-7} 
 &  & ODIN & \checkmark &  &  & \cite{cao2020} \cite{berger2021} \cite{araujo2023} \cite{calli2022} \cite{kim2022} \\ \cline{3-7} 
 &  & MC-dropout & \checkmark &  &  & \cite{berger2021} \cite{araujo2023} \cite{thagaard2020} \cite{linmans2020a} \cite{linmans2023b}  \cite{tardy2019} \cite{combalia2020} \\ \cline{3-7} 
 &  & Test-time Augmentation & \checkmark &  &  & \cite{araujo2023} \cite{combalia2020} \\ \cline{3-7} 
 &  & Deep Ensemble &  & \checkmark &  & \cite{berger2021} \cite{araujo2023} \cite{thagaard2020} \cite{linmans2020a} \cite{linmans2023b} \\ \cline{3-7} 
 &  & M-head CNN &  & \checkmark &  & \cite{linmans2020a} \cite{linmans2023b} \\ \cline{2-7} 

 & \multirow{2}{*}{\begin{tabular}[c]{@{}c@{}}Learning-based \\ deterministic UQ\end{tabular}} 
 & Evidential Deep Learning &  & \checkmark &  & \cite{araujo2023} \cite{tardy2019} \\ \cline{3-7}
 &  & Confidence Branch &  & \checkmark &  & \cite{zhang2021} \\ \cline{2-7} 
 
 & \multirow{3}{*}{\begin{tabular}[c]{@{}c@{}}OOD-aware\\  training\end{tabular}} 
 & Outlier Exposure &  & \checkmark &  & \cite{zhang2018} \cite{araujo2023} \cite{roy2022} \\ \cline{3-7} 
 &  & Reject Bucket &  & \checkmark &  & \cite{araujo2023} \cite{roy2022} \\ \cline{3-7} 
 &  & Dirichlet Prior Network &  & \checkmark &  & \cite{nandy2021} \\ \cline{2-7}
 
 & \multirow{3}{*}{\begin{tabular}[c]{@{}c@{}}Unsupervised \\ stand-alone detectors\end{tabular}} 
 & Autoencoder &  &  & \checkmark & \cite{cao2020} \\ \cline{3-7} 
 &  & Variational Autoencoder &  &  & \checkmark & \cite{cao2020} \\ \cline{3-7} 
 &  & Diffusion Models &  &  & \checkmark & \cite{graham2023a} \\ \hline

\multirow{13}{*}{\begin{tabular}[c]{@{}c@{}}Medical image \\ segmentation\end{tabular}} 

 & \multirow{2}{*}{\begin{tabular}[c]{@{}c@{}}Post-hoc \\ feature process\end{tabular}}  
 & Mahanlanobis & \checkmark &  &  & \cite{gonzalez2021} \cite{gonzalez2022} \cite{woodland2023} \cite{karimi2022} \\ \cline{3-7} 
 &  & Spectrum Decomposition & \checkmark &  &  & \cite{karimi2022} \\ \cline{2-7} 
 
 & \multirow{7}{*}{\begin{tabular}[c]{@{}c@{}}Learning-free\\  UQ\end{tabular}} 
  & MSP & \checkmark &  &  & \cite{gonzalez2021} \cite{gonzalez2022} \cite{yang2024} \\ \cline{3-7} 
  &  & Entropy & \checkmark &  &  & \cite{mehrtash2020} \\ \cline{3-7} 
  &  & KL from Uniform & \checkmark &  &  & \cite{gonzalez2022} \\ \cline{3-7} 
 &  & Temperature Scaling & \checkmark &  &  & \cite{gonzalez2021} \cite{gonzalez2022} \\ \cline{3-7} 
 &  & MC-dropout & \checkmark &  &  & \cite{gonzalez2021} \cite{gonzalez2022} \cite{karimi2022} \cite{yang2024} \\ \cline{3-7} 
 &  & Test-time Augmentation & \checkmark &  &  & \cite{gonzalez2022} \\ \cline{3-7} 
 &  & Deep Ensemble &  & \checkmark &  & \cite{karimi2022} \cite{yang2024} \\ \cline{2-7} 
 
 & \multirow{2}{*}{\begin{tabular}[c]{@{}c@{}}OOD-aware \\ training\end{tabular}}
 & EDL+RL &  & \checkmark &  & \cite{yang2024} \\ \cline{3-7} 
 &  & Outlier Exposure &  & \checkmark &  & \cite{karimi2022} \\ \cline{2-7}
 
 & \multirow{2}{*}{\begin{tabular}[c]{@{}c@{}}Unsupervised \\ stand-alone detectors\end{tabular}} 
 & Density Estimation &  &  & \checkmark & \cite{graham2022} \\ \cline{3-7} 
 &  & Latent Diffusion Models &  &  & \checkmark & \cite{graham2023b} \\ \hline

\end{tabular}
\label{table:3}
\end{table*}

\section{OOD detection in supervised medical image classification}
Research about OOD detection in medical image analysis mostly focuses on supervised medical image classification. While most adapted general OOD detection methods to medical fields, several techniques were first proposed to tackle specific clinical problems. Following the solution taxonomy described in \ref{sbusec:taxonomy}, we review the related studies in each category. Each subsection is organised as follows: the principle of the involved technique is first introduced and then followed by its application in medical image analysis.

\subsection{Post-hoc feature process}
One of these methods is feature-based binary classifier. A simple classifier, such as SVM, logistic regression, or KNN, is fitted by distinguishing between in-distribution and OOD samples in a validation set, where the input is the low-dimensional penultimate layer features extracted by base task networks. Feature-based binary classifier was evaluated in \cite{cao2020}, where medical OOD detection is benchmarked by comparing a variety of OOD methods across several medical image domains, including (1) chest X-ray image, (2) fundus images, and (3) stained histology slides of lymph nodes. Besides, all three distributional types are considered in their evaluation settings. Surprisingly, the logistic binary classifier outperformed all the other methods in the results that aggregate all evaluations, despite its simplicity.\\

Another representative method is Mahalanobis-based method, which was initially proposed by \cite{lee2018} to detect OOD input for a classification task. Mahalanobis distance between a data point $\mathbf{x}$ and a distribution with mean $\mu$ and covariance matrix $\mathbf{\Sigma}$ is defined as:
$$D_m = \sqrt{(\mathbf{x}-\mu)^T\cdot\mathbf{\Sigma}^{-1}\cdot(\mathbf{x}-\mu)}$$
Through multiplication with the inverse of $\mathbf{\Sigma}$, Mahalanobis distance rescales $\mathbf{x}$ into a covariance-free space where the outlier degree can be estimated more reasonably. Besides, it can also be viewed as a monotonic function of log-likelihood of Multivariate Gaussian Distribution, with a large $D_m$ indicating low density. \cite{lee2018} models the layer-wise feature of training samples as class-conditional Gaussian Distribution with tied covariance. For layer ${\ell}$, the feature map is global-average-pooled into a vector $\mathbf{x}$, then the confidence score $M_{\ell}(\mathbf{x})$ is defined as the negated squared $D_m$ between the test sample and closet class centroid:
$$M_{\ell}(\mathbf{x})=\max _c-\left(f_{\ell}(\mathbf{x})-\widehat{\mu}_{\ell, c}\right)^{\top} \widehat{\boldsymbol{\Sigma}}_{\ell}^{-1}\left(f_{\ell}(\mathbf{x})-\widehat{\mu}_{\ell, c}\right)$$
where $\widehat{\boldsymbol{\Sigma}}_{\ell}^{-1}$ and $\widehat{\mu}_{\ell, c}$ is the empirical covariance and empirical mean for class $c$ estimated over training set. To further improve OOD detection performance, the input is preprocessed by adding a perturbation  in gradient-sign direction similar to ODIN \cite{liang2017}:$$\widehat{\mathbf{x}}=\mathbf{x}-\varepsilon \operatorname{sign}\left(\nabla_{\mathbf{x}}\left(f_{\ell}(\mathbf{x})-\widehat{\mu}_{\ell, \widehat{c}}\right)^{\top} \widehat{\boldsymbol{\Sigma}}_{\ell}^{-1}\left(f_{\ell}(\mathbf{x})-\widehat{\mu}_{\ell, \widehat{c}}\right)\right)$$
Finally, the weighted sum over all layers is used to detect OOD, where the weights are estimated by fitting a logistic regression on the validation set.\\

\cite{berger2021} compared a couple of OOD methods in semantic shift detection across natural images and chest X-ray image, noting that the performance of Mahalanobis-based method drops sharply in the latter. The author attributes this to the less separation among classes for X-ray images than for natural images, which is substantiated by the T-SNE of intermediate layer features. Besdies, it is commonly observed that Mahalanobis-based method tends to perform pretty well in contextual shift detection but poorly in semantic shift detection. In \cite{cao2020}, it is demonstrated to be quite effective for all OOD types except for novel disease classes. A similar trend is also shown in OCT (Optical coherence tomography)-based retina disease classification \cite{araujo2023} and skin disease classification \cite{li2022}\cite{pacheco2020}\cite{roy2022}, where Mahalanobis-based method is used to compare with their proposed approaches. \cite{calli2022} proved Mahalanobis-based method with some modifications \cite{lee2018} performed pretty well in covariate shift detection of chest X-ray (CXR). They argued that Mahalanobis distance computed on the convolutional layer is linearly increased with the number of channels, thereby dividing the layer-wise score by this number to prevent the deep layer from weighing more. Besides, they estimate the mean of each class and common covariance using a validation set (only in-distribution) instead of the training set. In their experiments, the base task model is trained on posteroanterior (front to back) adult CXR, with the anteroposterior (back to front), lateral, pediatric CXR, and Non-CXR being OOD samples for evaluation.\\

Cosine similarity is also utilized to detect OOD in feature space \cite{araujo2023}. For two vectors $x$, $y$, it is given by $$s(x,y) = cos(\theta) = \frac{x\cdot y}{\lVert x \Vert_2  \lVert y \Vert_2}$$ which only measures the difference in direction without considering magnitude. Similar to \cite{lee2018}, the class centroid is empirically estimated over the penultimate layer representations of training samples. Given an input with its representation $x$, the OOD score is defined as below \cite{araujo2023}:$$S_{ood}=1-max\{s(x,c_1),...,s(x,c_K)\}$$ where $c_i$ is the centroid of class $c$. \cite{araujo2023} explored the semantic shift detection and contextual shift detection in OCT-based retina disease classification, with novel retina disease types absent in the training set and fundus images being corresponding OOD instances. They compared a series of prevalent OOD methods and metrics, finding that all methods can be significantly improved by simply using cosine similarity as a metric.\\

\cite{li2022} applied a popular Outlier Detection algorithm, Isolation Forests (IF), to the intermediate layer features of a trained CNN classifier, named Deep Isolation Forests. An Isolation Forest consists of multiple independent decision trees, with each of them constructed by iteratively splitting the nodes with randomly sampled feature and split point. The normality is then defined as:$$N(x) = -2^{-\frac{E[p]}{p_{avg}}}+0.5$$ where $E[p]$ is the mean of path length that traversal in each tree, $p_{avg}$ is the average path length over the training set. The intuition is an outlier may contain extreme feature value that deviates from normal samples and therefore can be easily isolated by a decision tree at the early stage. Taking a similar strategy to \cite{araujo2023}, \cite{li2022} constructed an IF for each class and $max\{IF_1,..,IF_C\}$ is used as In-distribution score. The evaluation is conducted eight times on semantic shift detection in skin lesion classification, with each lesion class being OOD in turn. Compared with some popular OOD detection methods, Deep Isolation Forests performed best in five of eight runs, suggesting it is a promising method for semantic shift detection in medical image classification.
 \\

Extreme Value Theorem (EVT) is first applied to open set recognition by \cite{bendale2016}. \cite{yasin2020} considered this method to equip a CNN-based skin disease classifier. After training, the penultimate activations (i.e., logits) are first extracted.  For any pre-defined class $c$, the mean activation vector $\mu_c$ is estimated over correctly classified training samples, and then a Weibull distribution is fitted on the largest distance between $\mu_c$ and the associated samples to estimate the probability of input $x$ being an outlier with respect to class $c$, noted as $w_c(x)$. During inference, OpenMax redistributes the logits and forms a new logit for the rejection (OOD) class. Specifically, the rejection logit $z_0$ is the weighted sum over the top $\alpha$ highest logits $\{z_1,...,z_\alpha\}$: 
$$z_0 = \sum_{i=1}^\alpha w_i(x)z_i$$ where the weight $w_i(x)$ is estimated by the per-class Weibull distribution. Besides, the $z_i$ is scaled to keep the total logits unchanged: $$z_i = (1-w_i(x))z_i$$ Finally, the probability of being OOD or each pre-defined class is explicitly output by performing SoftMax over $\{z_0, z_1,...,z_C\}$. The author \cite{yasin2020} simply evaluated this method on 10  images containing novel types of skin diseases absent in the training set (i.e., semantic shift detection) and 10 natural images (i.e., contextual shift detection), observing 80\% and 100\% OOD  samples are detected in two situations, respectively.\\

\cite{pacheco2020} focused on skin disease classification, evaluating the Gram matrix-based method proposed by \cite{sastry2020} in several OOD detection settings, where the healthy skin image, the corrupted skin image, and the natural image/histology image are OOD samples, respectively. For an image $D$, the activation map $A_{H\times W\times K}$ from the $l_{th}$ layer can be represented as a matrix $F_l=[v_1,...,v_K]^T$, where $v_{th}$ is the flattened feature feature map of the $k_{th}$ channel. Then $p-order$ gram matrix of the $l_{th}$ layer is given by:$$G_l^p=(F_l^p (F_l^p)^T)^{\frac{1}{p}}=[(g_{ij})_l^p]_{K\times K}$$ where $(g_{ij})_l^p = <v_i^p,v_j^p>$ encodes the correlation between feature map pairs and $v_i^p$ is derived by taking the power $p$ of $v_i$, with higher order focusing more on the prominent activations. After the classification model training, the upper bound (i.e., max) and lower bound (i.e., min) of each $g_{ij}^p $ are estimated across the training set. During inference, the deviation from the training set interval is computed for each order and each layer, with their aggregation viewed as a signal of abnormality to detect OOD. The experiments demonstrate that the Gram matrix-based method performed better than Mahalanobis-based method in an unbiased evaluation setting, where the validation set (containing both in-distribution and OOD samples) is unavailable for hyperparameter-tuning.\\

\cite{kim2022} also explored skin disease classification and proposed to apply subset scanning \cite{cintas2021} to OOD detection. Given a trained model, an input, and a layer, they search a subset of all nodes on the layer, where the divergence between input’s activations and in-distribution activations is maximized. The anomalousness is then quantified by a log-likelihood ratio statistic (e.g., the Berk-Jones test statistic), and the sum over all layers is thresholded to detect OOD. Besides, the author also found the ODIN perturbations \cite{liang2017} further improved this method. The evaluation is conducted on both semantic shift detection and covariate shift detection, where unseen skin disease images and skin disease images collected with different acquisition protocols are used, respectively. Although subset scanning is most effective in covariate shift detection as compared to MSP and ODIN, it is even inferior to MSP in semantic shift detection.

\subsection{Learning-free uncertainty quantification}
A basic learning-free UQ method is Maximum Softmax Probability (MSP) \cite{hendrycks2016}, which simplifies the model posterior ${\mathrm{p}(\boldsymbol{\theta} \mid \mathcal{D})}$ as a single point estimate. A threshold is determined based on the validation set, and the input with MSP lower than the threshold is detected as OOD. Despite underperformance, MSP is widely used as the baseline in a range of studies\cite{cao2020}\cite{zhang2021}\cite{berger2021}\cite{araujo2023}\cite{linmans2023b}\cite{calli2022}\cite{kim2022}\cite{roy2022} due to its simplicity. Besides, a variant is also used in \cite{linmans2020a} \cite{linmans2023b}, where the uncertainty, or OOD-ness, is quantified by the entropy of softmax probability. \cite{guo2017} suggested calibrating the prediction through temperature scaling. When training is done, softmax score is rescaled by divining a coefficient $T$ from logits vector $f(\boldsymbol{x})$:
$$S_i(\boldsymbol{x} ; T)=\frac{\exp \left(f_i(\boldsymbol{x}) / T\right)}{\sum_{j=1}^N \exp \left(f_j(\boldsymbol{x}) / T\right)}$$ where temperature $T$ is tuned on the validation set. In this way, only the magnitude of MSP is scaled while the predicted class remains unchanged. Due to implementation simplicity, it is commonly evaluated in medical image OOD detection \cite{berger2021} \cite{araujo2023} \cite{linmans2023b} for comparison.\\

Later, ODIN \cite{liang2017}  is proposed to improve MSP so that it is more distinguishable between in-distribution and OOD samples. After model training, the input is first added a perturbation in the opposite way of fast gradient sign method (FGSM)\cite{goodfellow2014}: $$\tilde{\boldsymbol{x}}=\boldsymbol{x}-\varepsilon \operatorname{sign}\left(-\nabla_{\boldsymbol{x}} \log S_{\hat{y}}(\boldsymbol{x} ; T)\right)$$ where $\varepsilon $ is the magnitude and $ S_{\hat{y}}(\boldsymbol{x} ; T)$ is the original MSP with temperature $T$. Then the MSP of new input $\tilde{\boldsymbol{x}}$ is computed and thresholded to detect OOD samples. The temperature, $T$, perturbation magnitude $\varepsilon$, and threshold $\sigma$ are tuned on a validation set containing both in-distribution samples and OOD samples to reach a 95\% TPR (i.e., keeping the MSP of 95\% of in-distribution samples higher than the threshold $\sigma$). Moving the input in a fast gradient sign direction w.r.t MSP, the perturbation inflates the score to be higher. Further, the effect is experimentally shown to be more influential to in-distribution samples than to OOD, which promotes a larger gap between them. ODIN is also broadly evaluated \cite{cao2020} \cite{berger2021} \cite{araujo2023} \cite{calli2022}\cite{kim2022} for OOD detection in medical image classification. \cite{berger2021} experimentally demonstrated ODIN is the more effective than Mahalanobis-based method, MC-Dropout, and Deep Ensemble on semantic shift detection in lung X-ray pathology classification. Besides, they also concluded that the improvement mainly comes from perturbation instead of temperature scaling. \cite{araujo2023} found that replacing MSP with cosine similarity dramatically improved the performance of ODIN on both semantic shift detection and contextual shift detection.\\

Dropout \cite{srivastava2014} is a popular regularization technique for deep neural networks, which randomly zeros out a fraction of layer nodes during training to alleviate over-fitting. Another UQ method prevalently used for OOD detection is MC-Dropout \cite{gal2016}, which utilizes dropout during inference to generate a set of predictions.  Given an test sample $\boldsymbol{x}^*$, MC-Dropout approximates the distribution over all possible predictions through a set of point estimates $\{\mathrm{P_i}\left(y=\omega_c \mid \boldsymbol{x}^*, \boldsymbol{\theta}_i\right)\}_{i=1,...,T}$ generated by applying $T$ randomly sampled dropout masks. Then the expected prediction for the class $c$ is estimated by the sample mean:
$$\mathrm{E}_{p\left(y=\omega_c \mid \boldsymbol{x}^*, \mathcal{D}\right)}\approx
\hat{y_c}=\frac{1}{T}\sum_{i=1,...,T} \mathrm{P_i}\left(y=\omega_c \mid \boldsymbol{x}^*, \boldsymbol{\theta}_i\right)$$ and the variance for the class $c$ is estimated by the sample variance:
$$\mathrm{Var}_{p\left(y=\omega_c \mid \boldsymbol{x}^*, \mathcal{D}\right)}\approx
\hat{\sigma_c}=\frac{1}{T}\sum_{i=1,...,T}[\mathrm{P_i}\left(y=\omega_c \mid \boldsymbol{x}^*, \boldsymbol{\theta}_i\right)-\hat{y_c}]^2$$
In \cite{berger2021}\cite{araujo2023}\cite{linmans2023b}\cite{linmans2020a}\cite{thagaard2020}\cite{tardy2019} where MC-dropout is evaluated for OOD detection upon medical image classification, three OOD scores are often used, including the MSP of expected prediction: $$MSP = max\{\hat{y_1},...,\hat{y_C}\}$$ the entropy of expected prediction: $$H_c=-\sum_{c=1} ^C \hat{y_c}\mathrm{ln}\hat{y_c}$$ and the average of class variance: $$\mathrm{Var} =\frac{1}{C} \sum_{i=1}^C \hat{\sigma_c}$$
\cite{berger2021} evaluate MC-dropout in semantic shift detection, with the MSP of expected prediction being the in-distribution score. The experiments reflect it is effective for natural images but unsatisfactory for chest X-ray image. \cite{araujo2023} evaluated MC-dropout with all three score functions, demonstrating all of them are significantly surpassed by the cosine similarity-based method on both semantic and contextual shift detection in OCT-based retina disease classification. \cite{thagaard2020} evaluated MC-dropout in semantic shift detection on stained histology slides. In their setting, the base task is the detection of adenocarcinoma in hematoxylin and eosin (H\&E) lymph node sections from breast cancer, with squamous cell carcinoma (SCC) from head and neck cancer being OOD. The performance of MC-dropout is unsatisfactory and even worse than the baseline (MSP) in all evaluation metrics.\\

Test-time augmentation (TTAUG) is also a simple strategy to mimic distribution over predictions, which utilizes a series of augmentations $\{ T_i \}_{i=1,...,M}$ to generate different versions of input and feed them into the trained model to get a set of point estimates $\{\mathrm{P_i}\left(y=\omega_c \mid \boldsymbol{x}^*, \boldsymbol{\theta},T_i\right)\}_{i=1,...,M}$. \cite{araujo2023} also evaluated this method with the same OOD score as MC-dropout. The results show it is reliable for contextual shift detection but unsatisfactory for semantic shift detection. \cite{combalia2020} suggested a combination of TTAUG and MC dropout. They first process the test sample with different augmentations and then forward each version with a sampled dropout mask. Besides, a novel metric named Bhattacharyya Coefficient (BC) \cite{van2019} is considered, which measures the overlap between the prediction distribution (i.e., $\{\mathrm{P_i}\left(y=\omega_c \mid \boldsymbol{x}^*, \boldsymbol{\theta},T_i\right)\}_{i=1,...,M}$) of the top two classes with the highest expected prediction $\hat{y_c}$. The experiment is conducted on semantic shift detection in skin disease classification, demonstrating the combination is more effective than either using TTAUG or MC-dropout alone. Besides,  the mean of class variance is shown to be the optimal metric, superior to Bhattacharyya Coefficient.\\

Ensemble is a strategy that involves combining multiple base models to create a more accurate and robust model. Another way to approximate the distribution over predictions is the explicit networks ensemble, named Deep Ensemble (DE) \cite{lakshminarayanan2017}. During training,  a set of networks are trained in parallel:$$\boldsymbol{M} = \{M_{\boldsymbol{\theta}_1},...,M_{\boldsymbol{\theta}_T}\}$$ These networks share identical architecture but have random weight initializations and random training data shuffling to keep the variation among them. During inference, the input $\boldsymbol{x}^*$ is fed into all the models to obtain a set of predictions $\{\mathrm{P_i}\left(y=\omega_c \mid \boldsymbol{x}^*, M_{\boldsymbol {\theta}_i}\right)\}_{i=1,...,T}$. Similar to MC-dropout, the expected prediction for class $c$ is estimated by the sample mean:$$\mathrm{E}_{p\left(y=\omega_c \mid \boldsymbol{x}^*, \mathcal{D}\right)}\approx
\hat{y_c} =
\frac{1}{T}\sum_{i=1,...,T} \mathrm{P_i}\left(y=\omega_c \mid \boldsymbol{x}^*, M_{\boldsymbol{\theta}_i}\right)$$ and the variance of prediction for class  is estimated by the sample variance: 
\begin{equation*}
\begin{aligned}
\resizebox{1.0\hsize}{!}{$\mathrm{Var}_{p\left(y=\omega_c \mid \boldsymbol{x}^*, \mathcal{D}\right)}
\approx \hat{\sigma_c}=\frac{1}{T}\sum_{i=1,...,T}[\mathrm{P_i}\left(y=\omega_c \mid \boldsymbol{x}^*, M_{\boldsymbol{\theta}_i}\right)-\hat{y_c}]^2$}  
\end{aligned}  
\end{equation*} 
Three OOD scores for MC-dropout are also commonly used for DE. With MSP of expected prediction being in-distribution score, DE is shown to perform slightly better than the baseline but significantly worse than ODIN on semantic shift detection in chest X-ray disease classification \cite{berger2021}. \cite{araujo2023} tried three scores for DE on semantic shift detection in OCT-based retina disease classification, finding all of them are less effective than ODIN and cosine similarity. \cite{thagaard2020}\cite{linmans2023b}\cite{linmans2020a} both test DE semantic shift detection about lymph nodes histology slides, with the first using MSP of expected predictions and the latter two using the entropy of expected predictions. As compared to the baseline (MSP), DE shows a significant superiority in cite{thagaard2020} but only a limited improvement in \cite{linmans2023b} and \cite{linmans2020a}.\\

A disadvantage of Deep Ensemble is the huge computational overhead caused by multiple independent training and inference runs. To address the issue, \cite{linmans2020a} proposed a variant named multi-head CNN. It consists of a CNN backbone followed by several randomly initialized output heads, which generate a set of predictions in a single pass while reducing the computational burden by sharing the weights in the early-stage layer. Besides, the author suggested a loss function called meta loss, which is defined as the weighted sum of the cross entropies over all heads:$$\mathcal{M}(g(x), y)=\sum_{m=1}^M \delta_m \mathcal{L}\left(g^m(x), y\right)$$ where $g^m(x)$ is the softmax output of $m_{th}$ head, $\mathcal{L}$ is the cross entropy loss. The weight of each head is determined in the following manner:$$\delta_m= \begin{cases}1-\epsilon & \text { if } m=\arg \min _i \mathcal{L}\left(g^i(x), y\right) \\ \frac{\epsilon}{M-1} & \text { else. }\end{cases}$$ where $\epsilon$ is a small value to assign the winning head (with the minimum loss) the largest weight. In this way, the most gradient signals are distributed to that head to encourage specialization and promote the diversity of the ensemble. \cite{linmans2020a} evaluated M-head CNN on semantic shift detection in stained histology slides-based breast cancer metastasis identification, where the novel class, diffuse large B-cell lymphoma, is treated as OOD. They show 10-head CNN outperforms baseline, MC-dropout, and standard Deep Ensemble by a large margin in FPR 95. \cite{linmans2023b} further evaluated this method in covariate shift detection and contextual shift detection about lymph node histology slides, with three settings being (in-distribution vs. OOD): (1) prostate biopsies without colorectal tissue vs. prostate biopsies containing colorectal tissue; (2) lymph node tissue vs. prostate biopsies; and (3) prostate biopsies vs. lymph node tissue. However, M-head CNN has no obvious advantage over other methods in these evaluations.

\subsection{Learning-based deterministic uncertainty quantification}
A classic method under this category is Evidential Deep Learning (EDL) \cite{sensoy2018}. EDL is inspired by Subjective Logic \cite{josang2016}, which formalizes the Dempster–Shafer Theory of Evidence (DST) with Dirichlet distribution. Similar to DPN \cite{malinin2018} \cite{malinin2019} \cite{nandy2020}, it explicitly models the distribution over all possible predictions as a Dirichlet distribution parameterized by concentration parameters $\boldsymbol{\alpha}=(\alpha_1,...,\alpha_K)$: 
$$\operatorname{Dir}(\boldsymbol{p} \mid \boldsymbol{\alpha})=\frac{1}{B(\boldsymbol{\alpha)}}
\prod_{i=1}^K p_i^{\alpha_i-1}$$ where $B(\alpha)$ is the K-dimensional multimodal beta function, and $\boldsymbol{p}$ is the vector lies on the probability simplex, satisfying $\sum_{i=1}^Kp_i=1$. Further, the Dirichlet distribution is explained from the view of DST. Denote the collected evidence supporting class $k$ as $e^k$, the $e^k$ is associated with concentration parameter $\alpha_k$ through $e^k=\alpha_k-1$ and then the total evidence is Dirichlet strength (also known as precision) $S=\sum_{i=1}^K e_k+1$. The belief mass assigned to class  is defined as the ratio between per-class evidence and total evidence, given as $b^k=\frac{e_k}{S}=\frac{\alpha_k-1}{S}$, while the uncertainty mass  is computed so that all masses sum up to one, obtaining 
$$u=\frac{K}{S}, u+\sum_{i=1}^K b_k=1$$ In this framework, a Dirichlet corresponds to a belief mass assignment (or, opinion), which is based on the evidence observed from data. Besides, the aleatoric uncertainty (AU) and epistemic uncertainty (EU) can be separately quantified as the mean of Dirichlet distribution, i.e., the expected categorical prediction:
$$\hat{p}_k=\frac{\alpha_k}{S}$$ and the (epistemic) uncertainty mass:
$$u=\frac{K}{S}$$
which is inversely proportional to the total amount of evidence. An observation without any evidence found corresponds to the maximum epistemic uncertainty $u=1$, while sufficient evidence could reduce the epistemic uncertainty $u$ to trivial.\\

Given a classification network, the softmax layer is replaced with a ReLU activation layer to generate the evidence for an instance $\mathbf{x}$, noted as $e_k = f_k(\mathbf{x}|\boldsymbol{\theta})$. Then a Dirichlet distribution $\operatorname{Dir}(\boldsymbol{p} \mid \boldsymbol{\alpha})$ is parameterized by $$\alpha_k = f_k(\mathbf{x}|\boldsymbol{\theta})+1$$, with each prediction $\boldsymbol{\hat{p}}=(\hat{p}_1,...,\hat{p}_K)$ being viewed as drawn from this distribution. The total evidence, or Dirichlet strength, is given by $S=\sum_{i=1}^K[f_i(\mathbf{x}|\boldsymbol{\theta})+1]$. However, the loss for $\mathbf{x}$ can not be directly computed as the the sampling process is undifferentiable. Alternatively, the expectation of loss with respect to the Dirichlet distribution $\operatorname{Dir}(\boldsymbol{\hat{p}} \mid \boldsymbol{\alpha})$ is used to train the network:
\begin{equation*}
\begin{aligned}
\mathcal{L}(\mathbf{x}, \boldsymbol{\theta}) & =\int\left\|\mathbf{y}-\boldsymbol{p}\right\|_2^2 \frac{1}{B\left(\boldsymbol{\alpha}\right)} \prod_{i=1}^K p_{i}^{\alpha_{i}-1} d \boldsymbol{p} \\
&
=\sum_{i=1}^K\left(y_{i}-\hat{p}_{i}\right)^2+\frac{\hat{p}_{i}\left(1-\hat{p}_{i}\right)}{\left(S+1\right)}
\end{aligned}
\end{equation*}
where  $\mathbf{y}=(y_{1},...,y_{K})$ is the label and $\boldsymbol{p}$ is the predicted probability vector, while $\boldsymbol{\hat{p}} = (\hat{p}_1,...,\hat{p}_K)$ is the expectation of $\boldsymbol{p}$, given by $\hat{p}_k=\frac{\alpha_k}{S}$. Besides, to prevent the evidence of incorrect class from being increased, a regularizer is added to the above loss: $$\mathcal{L}(\theta)= \mathcal{L}(\theta)+
\lambda_t K L\left[D\left(\boldsymbol{p} \mid \tilde{\boldsymbol{\alpha}}\right) \| D\left(\boldsymbol{p} \mid\langle 1, \ldots, 1\rangle\right)\right]$$ where $\tilde{\boldsymbol{\alpha}}=\mathbf{y}+\left(1-\mathbf{y}\right) \odot \boldsymbol{\alpha}$ and the weight $\lambda_t=min(1, \frac{t}{10})$ increasing gradually until the $10_{th}$ epoch. This is equivalent to forcing the evidence $e_{k \neq j} = \alpha_{k \neq j}-1$ to be zero except for the evidence supporting the correct class $j$. Training with this loss assures the uncertainty $u=\frac{K}{S}$ is reduced only when the evidence in favor of the ground truth class is found enough, or in other words, high uncertainty is reflected by the lack of correct evidence.\\ 

\cite{araujo2023} compared EDL with other methods on both semantic and contextual shift detection in OCT-based retina disease classification. The experiment showed its performance is quite terrible in both evaluations and even worse than the baseline (MSP) in semantic shift detection, suggesting the uncertainty estimation learned from the in-distribution cannot generalize well to OOD samples. \cite{tardy2019} directly applied the subjective logic-based UQ framework in EDL to a general pre-trained classification network, without replacing the softmax with ReLU and retraining with a special loss. Instead, they manually rescale the logits into a specified non-negative range to be the evidence and qualify the uncertainty with $\mu$. Besides, the author also suggested using Mahalanobis-based method as a complementary to further improve the OOD detection. They artificially generate a linear transition from the in-distribution sample, i.e., Full-Field Digital Mammography (FFDM) images, to the OOD sample, i.e., 2D views synthesized from 3D tomosynthesis acquisitions (S-View), in order to mimic a series of samples that gradually deviate from in-distribution. The uncertainty mass $\mu$ is observed to be most effective at the middle degree of transition and degenerates thereafter. In contrast, Mahalanobis distance is more indicative of OOD in the last half, substantiating the two are complementary with each other. In their experiment, three breast imaging classification tasks are considered: (1) risk assessment (high vs. low risk), (2) breast density stratification according to BI-RADS scores, and (3) glandular vs. conjunctive patch-tissue classification. However, the covariate shift detection is evaluated in a non-straightforward way. Rather than measuring the detection performance, they mix the covariate shiftsamples and normal samples, compare the base classification task performance when deleting the most uncertain samples with different threshold levels, and see if improvements are observed by doing so. The results show the proposed method is effective and comparable to EDL\cite{sensoy2018} and MC-dropout\cite{gal2016} while requires no retraining and model modification. \\

Another learning-based deterministic UQ method is the confidence branch \cite{devries2018}, which adds a branch parallel to the classification head to explicitly output the confidence estimation. Specifically, the confidence branch takes the global feature as input and outputs confidence estimation $c\in[0,1]$ through multiple fully connected layers followed by a sigmoid activation function. During training, the model is allowed to correct its classification prediction by asking for hints, which is achieved by interpolation between the prediction and the ground truth label: $$p_i^{\prime}=c \cdot p_i+(1-c) y_i$$ where confidence $c$ decides the degree to request a hint. To avoid the model lazily learning confidence $c$ to be a constant zero, the access to hints is penalized by a loss: $$\mathcal{L}_c=-\log (c)$$ Finally, it is added to the base task loss, leading to the total loss as below: $$\mathcal{L}=\mathcal{L}_t+\lambda \mathcal{L}_c$$ where $\lambda$ is a hyperparameter to balance the two losses. Minimizing this loss forces the model to access the hint only when it has no confidence about its prediction, which is equivalent to measuring the confidence by the model’s willingness to request hints. Further, four tricks are used to improve the methods. First, the $\lambda$ is adjusted dynamically through the training process, which is achieved by setting a fixed budget $\beta$ and increasing (decreasing) the $\lambda$ when $\mathcal{L}_c>\beta$ ($\mathcal{L}_c<\beta$) after each weight update. Then the access to ground truth, i.e., the interpolation, is adopted at only half of the batches, allowing the model to have a chance of 50\% to learn the classification from the error without answer disclosure. This can be explained as preventing the model from losing the ability to correctly classify. Besides, augmentations are used to create more hard-to-classify examples from which the pattern of low confidence is learned. Finally, the ODIN \cite{liang2017} is used w.r.t the hint loss $\mathcal{L}_c$, which is experimentally found to enlarge the gap between in-distribution and OOD samples.\\

\cite{zhang2021} utilized confidence branch to detect the contextual shifts in supervised medical image classification, evaluated the performance across multiple modalities and areas of concern, and compared to MSP and Outlier Exposure (OE). The results show it outperforms MSP by a large margin in all the evaluations while beating the OE in part of the evaluations as well, suggesting it is a promising solution to contextual shift detection. However, their performance on harder tasks, i.e., covariate shift detection and semantic shift detection, requires further exploration.

\subsection{OOD-aware Training}

Outlier Exposure (OE) was pioneered by \cite{hendrycks2018} to tackle OOD detection. They proposed to introduce OOD samples into the training set to heuristically learn the discrimination between In-distribution and OOD samples, in the hope that the effect can generalize to the unseen samples. In the case of classification, the author referred to \cite{lee2017}, using the cross-entropy from logits to uniform distribution to penalize the OOD samples. This can be interpreted as forcing the model to evenly distribute the propensities to all predefined classes for an OOD sample. As a result, no decision should be made based on the result. Thus, the final loss function is rewritten as below:$$\mathcal L = \mathbb{E}_{(x,y)\sim \mathcal{D}_{in}} \mathcal H_{ce}(y,f(x))+\mathbb{E}_{(x,y)\sim \mathcal{D}_{out}} \mathcal H_{ce}(U,f(x))$$ where $\mathcal H_{ce}$ is cross entropy and $U$ is uniform distribution. During inference, the entropy over the predicted probability vector $\boldsymbol{{\hat{p}}}=(\hat{p_1},...,\hat{p_K})$ is used as the OOD score. \cite{zhang2021} evaluated Outlier Exposure in contextual shift detection. Three datasets involving chest X-ray images, Musculoskeletal X-ray images (including elbow, finger, forearm, hand, humerus, shoulder and wrist), and fundus images are used as the in-distribution training set in turn, with the remaining two being test OOD samples. To simulate real clinical scenarios where the knowledge of possible OOD is incomplete, the exposed OOD instances (used for training) are only sampled from another hand X-ray dataset, with the unused instances being test OOD as well. Although OE outperforms MSP by a large margin, it slightly sacrifices classification accuracy as compared to the standard base task model.\\

The abstention class approach \cite{thulasidasan2020}, also known as the reject bucket \cite{araujo2023}, explicitly outputs the probability of being OOD through an extra class head, which is learned by introducing a few OOD samples as positive instances. \cite{araujo2023} explored few-shot OE and reject bucket, proving that just a small number of OOD exposures can aid semantic shift detection while maintaining the high accuracy of in-distribution retina disease classification. In addition, reject bucket is superior to standard OE in semantic shift detection in this case. \cite{roy2022} focus on semantic shift detection upon skin lesion classification, using a variant of reject bucket to detect the skin lesion classes beyond the training set. Specifically, they add several extra output heads, with each of them corresponding to a rare skin lesion class. Then a few instances of these rare classes are added to the original training set as exposed OOD samples. Besides, each training instance is associated with a fine-grain label specifying skin lesion class and a coarse-grain binary label indicating OOD or in-distribution. During training, the total loss is defined as the sum of fine-grain loss and coarse-grain loss:$$\mathcal L = \mathcal L_{fine}+\lambda \mathcal L_{coarse}$$ While $\mathcal L_{fine}$ is the normal cross entropy over skin lesion class labels, $\mathcal L_{coarse}$ is a binary cross entropy over coarse-grain labels, where the probability of being OOD is computed as the sum over all rare class heads. The experiment demonstrates it outperforms single reject bucket \cite{thulasidasan2020} by around 3 points in AUROC.\\

The idea of OE is also shown in Dirichlet Prior Network (DPN) \cite{malinin2018}, where the OOD training samples are used to model their behavior as distinct from in-distribution data. Recall most uncertainty quantification (UQ) methods approximate distribution over all possible predictions through a set of point estimates. For K-classification problem, each of the point estimates is a categorical distribution $\boldsymbol{\mu}=(\mu_1,...,\mu_K)$ over K-dimensional probability simplex. Given an input $\boldsymbol{x}^*$, DPN explicitly models such an ensemble by a parameterized Dirichlet distribution over all possible predictive categorical distributions: $$\mathrm{P}\left(y=\omega_c \mid \boldsymbol{x}^*, \mathcal{D}\right)=\int \underbrace{\mathrm{P}\left(y=\omega_c \mid \boldsymbol{\mu} \right)}_{\text {aleatoric }} \underbrace{\mathrm{p}(\boldsymbol{\mu} \mid \boldsymbol{x}^*,\mathcal{D})}_{\text {epistemic
 }} d \boldsymbol{\mu}$$ $$p(\boldsymbol{\mu}|\boldsymbol{x}^*,\boldsymbol{\theta})=\operatorname{Dir}(\boldsymbol{\mu} \mid \boldsymbol{\alpha})=\frac{\Gamma\left(\alpha_0\right)}{\prod_{c=1}^K \Gamma\left(\alpha_c\right)} \prod_{c=1}^K \mu_c^{\alpha_c-1}$$ where $\Gamma(\cdot)$ is the gamma function, $\boldsymbol{\alpha}=(\alpha_1,....,\alpha_K)$ is concentration parameter, and $\alpha_0 = \sum_{c=1}^K \alpha_c$ is precision controlling the sharpness of Dirichlet distribution. Thus, the final prediction is the expectation of Dirichlet distribution, given as: $$\mathrm{P}\left(y=\omega_c \mid \boldsymbol{x}^*, \mathcal{D}\right)=\frac{\alpha_c}{\alpha_0}$$ For an in-distribution input of class $c$, each point estimate of prediction, i.e., a categorical distribution $\boldsymbol{\mu}=(\mu_1,...,\mu_K)$, should have $\mu_c$ much larger than others. Besides, all the point estimates for this input should be consistent with each other. Thus, the Dirichlet distribution is expected to be significantly sharper at the corner (of probability simplex) associated with the ground truth. For those input of class $c$ but with high data uncertainty, each point estimate $\boldsymbol{\mu}=(\mu_1,...,\mu_K)$ may be flatter due to the inherent class overlap but all the point estimates should be still consistent, corresponding to a Dirichlet distribution sharper at the center. As for an OOD input, each point estimate $\boldsymbol{\mu}=(\mu_1,...,\mu_K)$ is flat while all the point estimates should be inconsistent with each other, which can be described by a Dirichlet distribution uniformly spread across the whole simplex. These behaviors can be modeled by minimizing Kullback-Leibler Divergence between the output Dirichlet distribution and a “template” with the expected nature, which is done in a multi-task fashion\cite{malinin2018}:
 \begin{equation*}
\begin{split}
 \mathcal{L}(\boldsymbol{\theta})=\mathbb{E}_{\mathcal{D}_{in}}[K L[\operatorname{Dir}(\boldsymbol{\mu} \mid \boldsymbol{\alpha}^{in}) \| \mathrm{p}(\boldsymbol{\mu} \mid \boldsymbol{x}^* ; \boldsymbol{\theta})]]\\
 +\mathbb{E}_{\mathcal{D}_{out}}[K L[\operatorname{Dir}(\boldsymbol{\mu} \mid \boldsymbol{\alpha}^{out}) \| \mathrm{p}(\boldsymbol{\mu} \mid \boldsymbol{x} ^*; \boldsymbol{\theta})]]
\end{split} 
 \end{equation*} Here, the OOD template $\operatorname{Dir}(\boldsymbol{\mu} \mid \boldsymbol{\alpha}_{out})$ sets all concentration parameters $\alpha_{k}^{out}=1$, while the template of in-distribution $\operatorname{Dir}(\boldsymbol{\mu} \mid \boldsymbol{\alpha}_{in})$ is shaped by the concentration parameter $ \boldsymbol{\alpha}_{in}=(\alpha_{k}^{in},...,\alpha_{K}^{in})$, where  
$$\alpha_{k}^{in}= \begin{cases}\beta+1 & \text { if } c=k \\ 1 & \text { if } c \neq k\end{cases}$$ and $\beta$ is a hyperparameter set to be large (e.g., 100). During inference, the measurement of Mutual Information (MI) is used to detect OOD, which isolates the epistemic uncertainty by removing aleatoric (data) uncertainty from the total uncertainty:$$MI=\underbrace{\mathcal{H}\left[\mathbb{E}_{\mathrm{p}\left(\boldsymbol{\mu} \mid \boldsymbol{x}^* ; \hat{\boldsymbol{\theta}}\right)}[\mathrm{P}(y \mid \boldsymbol{\mu})]\right]}_{\text {Total Uncertainty }}-\underbrace{\mathbb{E}_{\mathrm{p}\left(\boldsymbol{\mu} \mid \boldsymbol{x}^* ; \hat{\boldsymbol{\theta}}\right)}[\mathcal{H}[\mathrm{P}(y \mid \boldsymbol{\mu})]]}_{\text {Expected Data Uncertainty }}$$ Given a test input, the higher MI reflects a flatter Dirichlet distribution, thus indicating it is likely to be an OOD sample.\\

Later, \cite{malinin2019} suggested replacing the KL divergence with reverse KL divergence, while \cite{nandy2020} proposed an alternative loss function for DPN to more effectively differentiate the OOD sample from the in-distribution sample of high data uncertainty: 
 \begin{equation*}
    \resizebox{1.0\hsize}{!}{$\mathcal{L}_{i n}\left(\boldsymbol{\theta}, \lambda_{i n}\right)=
-\log p(y \mid \boldsymbol{x}, \boldsymbol{\theta})-\frac{\lambda_{i n}}{K} \sum_{c=1}^K \operatorname{sigmoid}\left(z_c(\boldsymbol{x})\right)$}
\end{equation*}
 \begin{equation*}
    \resizebox{1.0\hsize}{!}{$\mathcal{L}_{out}\left(\boldsymbol{\theta} ; \lambda_{{out }}\right)=\mathcal{H}_{c e}(\mathcal{U} ; p(y \mid \boldsymbol{x}, \boldsymbol{\theta}))-\frac{\lambda_{{out }}}{K} \sum_{c=1}^K \operatorname{sigmoid}\left(z_c(\boldsymbol{x})\right)$}
\end{equation*} where $z_c(\boldsymbol{x})$ is the logit associated with class $c$. For in-distribution data, the cross entropy $\mathcal{H}_{c e}$ is used to force the mean of Dirichlet Distribution, i.e., the expected prediction, to be consistent with the class label. For OOD sample, it is used to shape the expected prediction to be uniformly distributed over all classes. Besides, $\lambda_{i n}>0$ encourages a larger precision for in-distribution samples and $\lambda_{out}<0$ penalizes the precision for OOD samples, which can be seen from the following and noting sigmoid is a monotonically increasing function:
$$\alpha_0 = \sum_{c=1}^K \alpha_c
=\sum_{c=1}^K e^{z_c(\boldsymbol{x})}$$ Thus, minimizing the loss function results in an unimodal Dirichlet distribution for in-distribution and a multimodal  Dirichlet distribution for OOD, respectively. \cite{nandy2021} designed a diabetic retinopathy (DR) screening pipeline capable of covariate shift detection and contextual shift detection, which is achieved by a combination of two DPNs. During training, in-distribution training samples are identical for both DPN, while the instances from another retina image dataset and non-retinal images are OOD training samples, respectively. During inference, the first DPN outputs the DR screen prediction (DR or healthy), as well as identifies input suffering covariate shift, while the other directly rejects the images of no interest (i.e., non-retina images).

\subsection{Unsupervised stand-alone detectors}
A representative unsupervised stand-alone detector is the reconstruction error-based method. \cite{lyudchik2016} proposed to train an autoencoder (AE)  \cite{rumelhart1986} \cite{hinton2006} on normal data for anomaly detection. Specifically, an AE is used to compress the normal data into the low-dimensional latent space and then recover their original dimension to get a reconstruction. Then the difference between input and reconstruction, reconstruction error, is minimized to train an AE. Intuitively, reconstruction error can be used to measure the degree of abnormality as AE trained on normal data cannot capture unfamiliar patterns caused by the deviation, leading to low-quality reconstruction.\\

A variant is \cite{an2015}, which replaces the AE with variational autoencoder (VAE) \cite{kingma2013} to reconstruct the normal data and estimate the anomaly score via reconstruction probability. While most literature simply explains the principle of VAE-based anomaly detection as the intuition that deviation leads to a poor reconstruction, we try to give a more thorough explanation.\\

In short, VAE models the probability density of a sample $\boldsymbol{x}$ as a continuous form of Gaussian Mixture:  $$p(\boldsymbol{x})=\int p(\boldsymbol{z})p(\boldsymbol{x}\mid\boldsymbol{z}) d\boldsymbol{z}$$
which explains the generation of $\boldsymbol{x}$ as the following process: the latent variable $\boldsymbol{z}$ is first sampled from a 
 standard Gaussian distribution $p(\boldsymbol{z}):=N(\boldsymbol{z};\boldsymbol{0}, \boldsymbol{I})$, while the $\boldsymbol{x}$ is then sampled from a Gaussian distribution $p(\boldsymbol{x}|\boldsymbol{z}):=N(\boldsymbol{x};\boldsymbol{\mu}_z, \boldsymbol{\sigma}_z^2)$ determined by $\boldsymbol{z}$. Given a training set, the density estimation can be achieved via maximum likelihood estimation (MLE). The logarithm likelihood of an input $\boldsymbol{x}$ can be factorized as below:
\begin{center}
\resizebox{.9\hsize}{!}
{$
\begin{aligned}
\log p(\boldsymbol{x})&=\int q(\boldsymbol{z} \mid \boldsymbol{x}) \log p(\boldsymbol{z})d\boldsymbol{z}\\
&=\int q(\boldsymbol{z} \mid \boldsymbol{x}) \log(\frac{p(\boldsymbol{x}\mid\boldsymbol{z}) p(\boldsymbol{z})}{q(\boldsymbol{z} \mid \boldsymbol{x})} )d\boldsymbol{z}+KL[q(\boldsymbol{z} \mid \boldsymbol{x})\|p(\boldsymbol{z} \mid \boldsymbol{x})]
\end{aligned}
$}
\end{center}
where $q(\boldsymbol{z} | \boldsymbol{x})$ is an arbitrary probability density. As $KL[q(\boldsymbol{z} | \boldsymbol{x})\|p(\boldsymbol{z} | \boldsymbol{x})]\geq0$ always holds true, the first term is the variational lower bound of $\log p(\boldsymbol{x})$, noted as $L_b$. Thus, the objective of VAE, i.e., maximum likelihood estimation (MLE), can be achieved by directly maximizing $L_b$. Further, the variational lower bound can be factored into two parts: 
\begin{center}
\resizebox{.9\hsize}{!}
{$
\begin{aligned}
L_b&=\int q(\boldsymbol{z} \mid \boldsymbol{x}) \log(\frac{ p(\boldsymbol{z})}{q(\boldsymbol{z} \mid \boldsymbol{x})} )d\boldsymbol{z}+
\int q(\boldsymbol{z} \mid \boldsymbol{x}) p(\boldsymbol{x} \mid \boldsymbol{z})d\boldsymbol{z}\\
&=-KL[q(\boldsymbol{z} \mid \boldsymbol{x})\|p(\boldsymbol{z})]+\mathbb{E}_{q(\boldsymbol{z} \mid \boldsymbol{x})} \log p(\boldsymbol{x} \mid \boldsymbol{z})
\end{aligned}
$}
\end{center}
Given an training sample $\boldsymbol{x}^*$, VAE simulates $q(\boldsymbol{z} \mid \boldsymbol{x}^*)$ via a Gaussian distribution $N(\boldsymbol{z};\boldsymbol{\mu}, \boldsymbol{\sigma}^2)$ where mean $\boldsymbol{\mu}$ and variance $\boldsymbol{\sigma}^2$ is generated by the encoder $G_{\boldsymbol{\theta}}(\boldsymbol{x}^*)$. Then the first term of variational lower bound is derived as:
\begin{center}
\resizebox{.9\hsize}{!}
{$
\begin{aligned}
-KL[q(\boldsymbol{z} \mid \boldsymbol{x}^*)\|p(\boldsymbol{z})]=\frac{1}{2} \sum_{j=1}^J\left(1+\log \left(\left(\sigma_j\right)^2\right)-\left(\mu_j\right)^2-\left(\sigma_j\right)^2\right)
\end{aligned}
$}
\end{center}
where $J$ is the dimension of $\boldsymbol{z}$. Besides, a set of latent variables $\{\boldsymbol{z}_1,...,\boldsymbol{z}_M\}$ are sampled from $N(\boldsymbol{z};\boldsymbol{\mu}, \boldsymbol{\sigma}^2)$ and fed into the decoder $D_{\boldsymbol{\theta}}(\boldsymbol{z})$ to generate a set of mean and variance $\{\boldsymbol{(\mu}_1,\boldsymbol{\sigma}_1^2)
...,\boldsymbol{(\mu}_M,\boldsymbol{\sigma}_M^2)\}$, with each pair parameterized a Gaussian distribution $N_i(\boldsymbol{x};\boldsymbol{\mu}_i,\boldsymbol{\sigma}_i^2)$ associated with the sampled $\boldsymbol{z}_i$. Then the $N_i(\boldsymbol{x}^*;\boldsymbol{\mu}_i,\boldsymbol{\sigma}_i^2)$ is a mimic of $p(\boldsymbol{x}^* \mid \boldsymbol{z}_i)$ while the expectation is a mimic of the second term of variational lower bound:$$\mathbb{E}_{q(\boldsymbol{z} \mid \boldsymbol{x})} \log p(\boldsymbol{x}^* \mid \boldsymbol{z})=\sum_{i=1}^M \log N_i(\boldsymbol{x}^*;\boldsymbol{\mu}_i,\boldsymbol{\sigma}_i^2)$$ During training, $q(\boldsymbol{z} \mid \boldsymbol{x})$ gradually converge to the prior $p(\boldsymbol{z})$ as the Kullback-Leibler Divergence between them is minimized. Then the following equation can be seen as an approximation of the true density at the position of input $\boldsymbol{x}$:$$\hat{p}(\boldsymbol{x})\approx
\mathbb{E}_{q(\boldsymbol{z} \mid \boldsymbol{x})}  p(\boldsymbol{x} \mid \boldsymbol{z})=\sum_{i=1}^M N_i(\boldsymbol{x};\boldsymbol{\mu}_i,\boldsymbol{\sigma}_i^2)$$ which is termed reconstruction probability in \cite{an2015}. Thus, a low reconstruction probability reflects the input lies in the low-density region of normal data, which suggests a high possibility of being an anomaly. A simplification is using the expectation of the difference $\boldsymbol{\mu}_i-\boldsymbol{x}$, namely the reconstruction error of VAE, to be the anomaly score as $\boldsymbol{\mu}_i-\boldsymbol{x}$ is inversely proportional to $p(\boldsymbol{x} \mid \boldsymbol{z}_i)=N_i(\boldsymbol{x};\boldsymbol{\mu}_i,\boldsymbol{\sigma}_i^2)$.\\

As we mentioned in the preliminary, OOD detection is a special case of anomaly detection, where the “normal data” are the samples sharing the same distribution as the base task training set. Therefore, construction-based anomaly detection methods can be easily adapted to OOD detection by training an AE or VAE only on the in-distribution data. \cite{cao2020} evaluate both AE and VAE in all three OOD detection settings w.r.t. multi-class medical image classification. Overall, both methods perform well in contextual shift detection and covariate shift detection, while they lose as compared to the two post-hoc feature process methods, binary classifier, and Mahalanobis-based method. Besides, both of them fail to detect semantic shift, which is thought to be harder than the other two types. We speculate the reason as the natural heterogeneity in the multi-class training set makes it hard to capture the pattern of all pre-defined classes, resulting in an undesirable reconstruction even for in-distribution samples and thereby an ambiguous distinction from OOD samples. \\

However, the reconstruction quality of AE or VAE is closely related to the dimension of information bottleneck (e.g., latent space), which is selected before training and determined during inference, leaving a costly process to tune this key hyperparameter. To address this issue, Denoising Diffusion Probabilistic Models (DDPM) \cite{ho2020} were recently utilized to achieve anomaly detection \cite{wyatt2022} and OOD detection \cite{graham2023a} due to their capability to generate a set of reconstructions from diverse noise levels. The score function, i.e., reconstruction quality, can be measured by considering a range of bottleneck choices in a single inference run. For a given input $\boldsymbol{x}_0$, the forward process generates a series of $\boldsymbol{x}_t$ via iteratively adding the noise, which is also known as the diffusion process:$$\boldsymbol{x}_t=\sqrt{1-\beta_t} \boldsymbol{x}_{t-1}+\sqrt{\beta_t} z_t, \quad z_t \sim N(\boldsymbol{0}, \boldsymbol{I})$$ where $z_t$ follows standard Gaussian distribution and diffusion rate $\beta_t$ controls the variance of added noise. By simply setting  $\beta_t$ increases along with the forward step $t$, $\boldsymbol{x}_{t}$ converges to standard Gaussian $N(\boldsymbol{0}, \boldsymbol{I})$. It can be easily derived from the following expression:$$\bar{\alpha}_t = \prod_{i=1}^t1-\beta_i$$ $$\boldsymbol{x}_t=\sqrt{\bar{\alpha}_t} \boldsymbol{x}_0+\sqrt{1-\bar{\alpha}_t} \bar{z}_t, \quad \bar{z}_t \sim N(0, \boldsymbol{I})$$ where $\lim_{t\to \infty} \bar{\alpha}_t = 0$.\\

Then $p(\boldsymbol{x}_0)$ can be expressed as the integration over a chain generated by T-step diffusion process:$$p(\boldsymbol{x}_0)=\int_{\boldsymbol{x}_1:\boldsymbol{x}_T} 
p(\boldsymbol{x}_{0} \mid \boldsymbol{x}_{1})...
p(\boldsymbol{x}_{T-1} \mid \boldsymbol{x}_{T})
p(\boldsymbol{x}_T) d\boldsymbol{x}_1:\boldsymbol{x}_T$$ which explains the generation of $\boldsymbol{x}_0$ as a Markov process containing $T+1$ steps: $\boldsymbol{x}_T $ is firstly drawn from the standard Gaussian distribution $N(0, \boldsymbol{I})$, and the $\boldsymbol{x}_i$ is generated by denoising $\boldsymbol{x}_i$ at each step. Similar to VAE, DDPM is also trained via the maximization of the variational lower bound:$$L_b=\mathbb{E}_{q(\boldsymbol{x}_1:\boldsymbol{x}_T \mid \boldsymbol{x}_0)} \log(\frac{ p(\boldsymbol{x}_0:\boldsymbol{x}_1)}{q(\boldsymbol{x}_1:\boldsymbol{x}_T \mid \boldsymbol{x}_0)} )$$ which is achieved by denoising the instances generated by forward diffusion steps in practice, and the reconstruction error reflects the density at the input position for the same reason as VAE. \cite{graham2023a} trained DDPM only on the in-distribution data to detect OOD samples. During inference, they generated $N$ reconstructions for a test sample by denoising from $N$ randomly sampled steps, and the average of similarities between reconstruction and input is used as the input score. Specifically, the similarity metric is computed as the sum of MSE, and LPIPS \cite{zhang2018} which measures the distances in intermediate layer features. Besides, the faster sample strategy, PLMS sampler \cite{liu2022}, is adopted to speed up the inference. The author evaluated the method on the simplest OOD detection task in medical image classification, i.e., contextual shift detection across multiple modalities and organs (Hand X-ray, Abdomen CT, Chest X-ray, Chest CT, Breast MRI, and Head CT), finding it performs almost perfectly. However, the covariate and semantic shift detection in medical image classification remain unexplored in their work.\\ 

\section{OOD detection in medical image segmentation}
Medical image segmentation, as one of the pivot tasks in computer-aided diagnostics, was recently empowered by emerging deep learning models such as U-net \cite{ronneberger2015}. However, the available training samples for medical image segmentation, especially for those with 3D modalities such as CT and MRI, are quite rare due to the vast annotation cost. As a result, it is common to encounter distributional shift when applying the segmentation models to real clinical samples. The segmentation model trained on a specific dataset may silently output a meaningless or low-quality segmentation mask for the input from a different distribution. Recently, a range of research has paid attention to OOD detection in medical image segmentation. In this section, we also review them following the methodology taxonomy described in \ref{sbusec:taxonomy}. 

\subsection{Post-hoc Feature Process}

Similar to supervised medical image classification, most OOD detection methods in medical image segmentation distinguish between OOD sample and in-distribution sample based on the intermediate features extracted from the segmentation network instead of the final output. Mahalanobis-based method \cite{lee2018} is popularly utilized due to its applicability.\cite{gonzalez2021} focus on lung lesion segmentation in chest Computed Axial Tomography (CAT) scans of COVID-19 patients, finding that a patch-based nnU-net \cite{isensee2021}, even trained on a multi-center dataset, may still output an unreliable lesion mask on other datasets. Inspired by \cite{lee2018}, they extracted the features of all training patches from the encoder, downsampled them by average pooling, and then computed the mean and covariance to estimate a Gaussian Distribution. For the test sample, the low-dimension feature of each patch is extracted in the same way as above, and the Mahalanobis distance to the Gaussian Distribution, serving as the per-patch uncertainty estimation, is computed to finally combine into an image-level uncertainty mask in the same way as the prediction mask. Finally,  the average over all voxels is thresholded to determine OOD samples. In the experiment, datasets different from the training set in patient groups and acquisition protocols are viewed as OOD samples, and the result shows Mahalanobis-based method outperforms MSP, temperature scaling, and MC-dropout by a large margin in both Detection error and FPR. \cite{gonzalez2022} compared this method to others on a broader collection of OOD samples, including (1) training set data with artificial transformation (covariate shift); (2) chest CAT scans from patients suffering other lung diseases than COVID-19 (semantic shift), and (3) Spleen and Colon CT scans (contextual shift). Compared to other methods, Mahalanobis-based method stands the best across all evaluations. \cite{woodland2023} explored OOD detection for liver segmentation in T1-weighted liver magnetic resonance imaging exams (MRIs). Similar to \cite{gonzalez2022} and \cite{gonzalez2021}, they applied the Mahalanobis-based method on the bottleneck features of Swin UNETR \cite{hatamizadeh2021}. Besides, they explore four dimensionality reduction methods to reduce the feature dimension, including average pooling, PCA, UMAP \cite{mcinnes2018}, and t-SNE \cite{van2008}. In evaluation setting, the OOD samples come from either T1-weighted liver MRI that is hard to segment or T1-weighted liver MRI with poor image quality, which belongs to covariate shiftin our framework. The experiment substantiates that PCA with 256 principal components improves Mahalanobis-based method most.\\

\cite{karimi2022} suggested a multitask learning strategy for 3D organ segmentation to tackle the insufficiency of annotated training samples. Specifically, the model is trained on a mixture of CT scans and MRIs across multiple organs, including the brain cortical plate, liver, kidney, left atrium, prostate, pancreas, hippocampus, and spleen. Then they proposed to detect OOD samples using spectral analysis of the feature map extracted from the final convolutional layer. Given the feature map of size  denoting height, width, depth, and channel respectively,  the spectral decomposition is conducted on the flattened feature maps  of size :
$$ F=USV^T$$ where $S$ is a diagonal matrix containing the singular values of $F$, also known as spectrum. The normalized spectrum:$$s= \frac{diag(S)}{\lVert x \rVert_2}$$ is shown to be distinguishable between in-distribution and OOD samples, and the OOD score is finally computed by the Euclid distance from the test sample to its nearest neighbor within the training set. The author considers both contextual shift detection and covariate shift detection in evaluation settings  (in-distribution vs. OOD): (1) the training dataset mixing brain cortical plate-T2 MRI, prostate-MRI, heart-MRI, liver-CT, and liver-MRI vs. pancreas CT, spleen CT, and hippocampus MRI,; (2) brain cortical plate-T2 MRI from newborns vs. brain cortical plate-T2 MRI from the young and the older. Experiment results reveal spectrum is superior to other popular OOD methods in detection accuracy and AUROC, including MC-Dropout, Deep Ensemble, ODIN, Mahalanobis-based method, and Outlier Exposure.

\subsection{Learning-free uncertainty quantification }
The fact that segmentation can be viewed as pixel (voxel)-level classification naturally induces a simple way to estimate the uncertainty of the whole predicted segmentation mask. That is, the uncertainty of each pixel (voxel) is estimated in the same manner as image classification, while the image-level uncertainty is often represented through the aggregation over all pixels (voxels). In this way, traditional UQ methods can be directly used for a segmentation task.  
In related literature \cite{gonzalez2021} \cite{gonzalez2022} \cite{karimi2022}, it is very common to evaluate the learning-free UQ methods as a comparison to their proposed method, due to the simplicity of implementation. \cite{gonzalez2021} demonstrated MC-dropout is inferior to Mahalanobis-based method but outperforms other leraning-free UQ methods by a large margin, including MSP, temperature scaling, and a variant of MSP measuring the Kullback-Leibler Divergence from softmax score to uniform distribution \cite{hendrycks2019}\cite{lee2017}. On the basis of \cite{gonzalez2021}, \cite{gonzalez2022} adopted more evaluations to mimic all three distributional shifts, while adding test-time augmentation (TTAUG) into comparison. Although not as good as Mahalanobis-based method, TTAUG is superior to other UQ methods across all evaluations, suggesting it is a promising solution to OOD detection. \cite{mehrtash2020} found the average of per-pixel entropies w.r.t the foreground probability is able to effectively distinguish samples from two different prostate segmentation datasets, which can be seen as a mimic of the covariate shiftsetting. Besides, they also observed a negative correlation between this score and the Dice coefficients \cite{milletari2016}, concluding it can serve as a measurement of segmentation quality during inference, with high average entropy indicating a low-quality segmentation and thereby an OOD sample.

\subsection{OOD-aware training}

For OOD detection in medical image segmentation, it is also possible to model distinct behaviors for in-distribution and OOD samples by explicitly introducing OOD supervision. \cite{yang2024} suggested using a reinforcement learning strategy to tune an Evidential Deep Learning (EDL) \cite{sensoy2018} model. Specifically, a segmentation network is adapted to the EDL model by replacing the softmax layer with softplus function. In this way, the logit $f_k(x|\theta)$ is explained as the evidence supporting the corresponding class, which is further used to parameterize a Dirichlet distribution serving as the conjugate prior over the predicted categorical distribution $\boldsymbol{p}$: 
$$\operatorname{Dir}(\boldsymbol{p} \mid \boldsymbol{\alpha})=\frac{1}{B(\boldsymbol{\alpha)}}
\prod_{i=1}^K p_i^{\alpha_i-1}$$ 
where $\alpha_k = f_k(x|\theta)+1$. Then the model is trained by minimizing the expectation of cross entropy associated with each possible point estimate, which is computed by integrating out $\boldsymbol{p}$ over the Dirichlet. In this way, the aleatoric uncertainty (AU) and epistemic uncertainty (EU) can be separately expressed as the expected categorical prediction, $\hat{p}_k=\frac{\alpha_k}{\sum_{k=1}^K \alpha_k}$, and the total evidence, ${\sum_{k=1}^K \alpha_k}$, with the former being the uncertainty (confidence) estimation of in-distribution sample while the latter being the uncertainty estimation of OOD sample.\\

To further improve the uncertainty estimation, \cite{yang2024} proposed tuning a policy network ${\pi_\phi(x)}$ on the validation set, where the OOD samples are synthesized by corrupting the in-distribution samples. The policy network ${\pi_\phi(x)}$ is first initialized by the pre-trained EDL network $\pi_{\hat{\theta}}(x)$, and then optimized through maximizing a reinforcement learning objective:
$$\mathcal{J}(\phi)=\mathbb{E}_{(x, y)}\left[R(\mu, \hat{y}, y)-\beta \log \left(\frac{\pi_\phi(x)}{\pi_{\hat{\theta}}(x)}\right)\right]$$
The second term penalizes the policy network ${\pi_\phi(x)}$ from deviating far from the original pre-trained EDL network $\pi_{\hat{\theta}}(x)$, while the first term is a reward function that encourages some desirable behaviors for in-distribution and OOD samples. Specifically, the reward for in-distribution is defined as the negative logarithm of the calibration metric, Expected Calibration Error (ECE), which measures how well the confidence aligns with the practical predicted accuracy:
\begin{equation*}
\begin{aligned}
R(\mu, y, \hat{y})&=-\log \mathrm{ECE}\\
&= -\log\left(\sum_{m=1}^M \frac{\left|N_m\right|}{N}\left|\operatorname{acc}\left(N_m\right)-\mathbb{E}_{x\in N_m}\mu(x)\right|\right)
\end{aligned}
\end{equation*}
where $\mu$, $y$, and $\hat{y}$ denotes aleatoric uncertainty (AU) estimation, ground truth, and prediction, $N_m$ is the set of in-distribution samples falling into the $m_{th}$ bin, and $N$ is the total number of in-distribution samples. ECE groups the in-distribution samples into $M$ bins according to confidence estimation, computes the difference between confidence mean and accuracy within each bin, and finally averages over all bins. For OOD samples, the reward is the ratio of the epistemic uncertainty (EU) between themselves and their in-distribution counterparts (the version before corruption):
$$R(\mu, y, \hat{y})= \frac{\sum_{s\in ood}\mu(s)}{\sum_{s\in in}\mu(s)}$$ 
where $\mu$ is the epistemic uncertainty (EU) estimation, $s$ is the pixel, and the EU of the whole image is estimated via the aggregation over all pixels. It is obvious the maximization of such reward functions improves the calibration of the in-distribution sample while encouraging a high EU for OOD samples. In order to allow a more efficient tuning under the constraints of the second penalty term, the author suggested a fine-grained parameter update scheme. Specifically, the update step of each parameter is weighted by its importance to the model outputs, which can be computed by the diagonal element of the fisher information matrix \cite{kirkpatrick2017}. In their evaluation, two in-distribution tasks are carried out, including the segmentation of four different soft tissues in laparoscopic cholecystectomy images and the segmentation of the submucosal tissue, mucosa tissue, muscle tissue, and blood vessel in endoscopic surgical images. In evaluation, the in-distribution images are corrupted to mimic the situation of covariate shift. The experiments showed the proposed method obtains a significant improvement when compared with a series of UQ methods, such as MC-dropout \cite{gal2016}, Deep Ensemble \cite{lakshminarayanan2017}, and DUQ \cite{van2020}.

\subsection{Unsupervised stand-alone detectors}

Similar to the case of classification, an unsupervised stand-alone OOD detector for medical image segmentation is typically a generative model trained only on in-distribution samples, with training and inference isolated to the base task.\\

Rather than measuring reconstruction errors, another generative model-based approach is to directly estimate the likelihood of the input being in-distribution. Given a sequence, $\boldsymbol{X} = \{ \boldsymbol{x}_1,...,\boldsymbol{x}_N \}$, its probability can be factorized into a chain of conditional probabilities:$$p(\boldsymbol{X}) =
p(\boldsymbol{x}_1)
p(\boldsymbol{x}_2 | \boldsymbol{x}_{1})...p(\boldsymbol{x}_{N} | \boldsymbol{x}_{N-1})$$ Representing the 3D volumes with a sequence, [25] estimated the probability density of in-distribution volumes by multiplication of the conditional probabilities along this chain. The first thing is to learn an efficient compression, which can represent the input with a relatively short sequence while keeping useful information as much as possible. Inspired by the high-resolution image synethsis\cite{esser2021}, the author achieved this goal through VQGAN \cite{esser2021}, where a discrete codebook $\mathcal{Z}=\left\{\boldsymbol{z}_k \in \mathbb{R}^{n} \right\}_{k=1}^K$ is learned to encode the rich information of image constituents, and the 3D volumes are then represented as a sequence of entries drawn from the codebook. Specifically, the input volume $\boldsymbol{V} = [\boldsymbol{v}_{ijl} ]_{H\times W \times D} $ is first fed into an encoder $E(\cdot)$ to produce a feature map $\boldsymbol{X} = [\boldsymbol{x}_{ijl} ]_{h\times w \times d} $, which is further quantized by replacing each voxel $\boldsymbol{x}_{ijl}$ (i.e., patch for the whole volume) with its nearest codebook entry $\boldsymbol{z}_{\sigma(i,j,l)}$ in terms of the $L_2$ norm, obtaining a new spatial representation encoded by $\mathcal{Z}$: $$\boldsymbol{X}_z = [\boldsymbol{z}_{\sigma(i,j,l)} ]_{h\times w \times d} $$ where ${\sigma(i,j,l)}$ maps the index of encoder output $\boldsymbol{X}$ to the index of codebook $\mathcal{Z}$. Then a decoder $G(\cdot)$ is responsible for recovering the volume size from the quantized spatial representation $\boldsymbol{X}_z $, obtaining the reconstruction $\hat{\boldsymbol{V}}$. Besides, a discriminator $D(\cdot)$ is added to distinguish between real and reconstructions to push the limit of compression. Finally, the VQGAN is trained end-to-end in an adversarial manner with the total loss:
\begin{equation*}
 \begin{aligned}
\mathcal{Q}^*=\underset{E, G, \mathcal{Z}}{\arg \min } \max _D \mathbb{E}_{\boldsymbol{V} \sim p(\boldsymbol{V})} & {\left[\mathcal{L}_{\mathrm{VQ}}(E, G, \mathcal{Z})\right.} \\
& \left.+\lambda \mathcal{L}_{\mathrm{GAN}}(\{E, G, \mathcal{Z}\}, D)\right]
\end{aligned}   
\end{equation*}
where the first term computes the reconstruction error and the second computes the classification loss of the discriminator. In order to estimate the probability density of input, the spatial quantized representation $\boldsymbol{X}_z = [\boldsymbol{z}_{\sigma(i,j,l)} ]_{h\times w \times d} $ is flattened into a sequence $\boldsymbol{X}_z = \{\boldsymbol{z}_{r} \}_{r=1,...,h\times w \times d} $, and the likelihood of each patch, $\hat{p}(\boldsymbol{z}_{r} | \boldsymbol{z}_{i< r})$, is predicted in an autoregressive manner via a transformer \cite{vaswani2017} and is optimized via Maximum Likelihood Estimation (MLE). The author evaluated this method in two OOD detection settings w.r.t segmenting Intracerebral Haemorrhages (ICH) in head CT, with manually corrupted in-distribution ICH head CT and 3D volumes (including CT and MRI) of other organs respectively simulating covariate shiftand contextual shift. They show the proposed method performs well in all evaluations except for subtle corruption such as low-level noising and left-to-right flip. However, we argue this is not a concern as the model should be robust to these slight covariate shifts instead of rejection.\\

Despite the effectiveness of \cite{graham2022}, \cite{graham2023b} argued that transformer-based likelihood estimation suffers several disadvantages including sensitivity to compression level, high memory requirements, and disability to output an input-size OOD score map. To tackle the issues, they considered detecting OOD with the reconstruction error of a DDPM trained only on in-distribution samples. As directly applying DDPM for 3D volume leads to dramatically increased computational overhead, the author proposed Latent Diffusion Models (LDMs), which conducts the forward process, i.e., adding noises, and reconstructions, i.e., denoising, on the downsampled latent representation compressed by VQGAN [88]. During inference, the denoised representation is then decoded to recover the input size, and the reconstruction errors (or, similarity) are computed in the original input space instead of downsampled feature space. The experiment is conducted in the same way as [25], and the results showed LDMs outperforms \cite{graham2022} in both contextual and covariate shift detection. Besides, the proposed method successfully addresses the three disadvantages of \cite{graham2022}, suggesting it is a promising solution to OOD detection especially for high-resolution 3D volumes.

\section{Evaluation protocols}

Given a medical image classification or segmentation task, the training, hyperparameter tuning, and evaluation of the model are conducted on three distinct parts of in-distribution dataset, $D_{in}^{train}$, $D_{in}^{val}$, and $D_{in}^{test}$. When considering OOD detection, the test set should consist of both held-out parts from the in-distribution dataset and the datasets standing for OOD samples, namely $D_{out}^{test}$, read $$D_{ood}^{test}=D_{in}^{test} \cup D_{out}^{test}$$
In addition, some OOD samples should be added to the validation set in order to determine score threshold and other hyperparameters, read
$$D_{ood}^{val}=D_{in}^{val} \cup D_{out}^{val}$$ 
Note it is more reasonable to maintain the difference between $D_{out}^{val}$ and $D_{out}^{test}$ to simulate the realistic clinic scenarios where OOD samples are hard to expect.\\

In evaluation, the contextual shift is often simulated by non-medical image datasets, such as ImageNet, or medical images with different \textbf{modality} and/or \textbf{area of concern} from in-distribution samples, while the semantic shiftis naturally represented by the medical images within the in-distribution context but containing \textbf{class of target} such as diseases/lesions of novel classes. For covariate shift, the \textbf{image quality} issues and differences in \textbf{acquisition protocols\textbackslash pre-processing} are conventionally mimicked by corrupting or transforming the in-distribution instances, while the different \textbf{imaging view} and \textbf{subject groups} are tested using corresponding instances. \\

Labeling the OOD as positive and in-distribution negative, the performance of an OOD detection method can be measured through binary classification evaluation metrics, including accuracy, detection errors, sensitivity (TPR), specificity (TNR), AUROC, AUPR, and FPR95. Note the first four involve a fixed detection threshold, which is determined by keeping most (typically 95\%) of the in-distribution validation instances correctly identified. \cite{yang2024} evaluated their method with two brand-new metrics, Pixel Ratio, and Box Ratio \cite{zepf2023}. Specifically, Pixel Ratio is defined as the ratio between the score of a manually corrupted version (randomly adding a white noise box to an image, i.e., a mimic of covariate shift) and its in-distribution counterpart, where the score is computed by aggregating the score of all pixels. Box Ratio is defined in the same way, except that the score is the sum of the pixels located within the corrupted box. In addition, the evaluation of the OOD detection method should also take the base task evaluation into account so that accurate detection is not achieved at the cost of sacrificing the model performance. Besides, \cite{tardy2019} adopted an indirect way to evaluate covariate shift detection. In each run, they kept a specified proportion of samples with the lowest uncertainty estimations and tested the base task performance over them. Ideally, an improvement should be observed when the proportion decreases, which demonstrates the real uncertain samples, or the covariate shiftsamples, are successfully scored higher than those certain ones. \\

\section{Challenges and future directions}

The existing research has shown success in contextual and covariate shift detection, while methods for semantic shift detection in supervised medical image classification remain underperformance across multiple modalities and medical departments. \cite{cao2020} evaluated a range of OOD detection methods, reporting a random-guess level accuracy and AUROC for all of them on semantic shift detection about front-view chest X-ray image. \cite{berger2021} observed a significant drop in semantic shift detection performance from natural image recognition to the disease recognition of chest X-ray image. \cite{roy2022} showed the best approach in unseen skin disease detection achieving around 80\% in AUROC. The experiments in \cite{li2022} also reflected an undesirable result for semantic shift detection in skin disease classification, with the AUROC of their proposed method varying from ~63\% to ~76\% as the test OOD disease changes. Similarly, \cite{kim2022} reported the best performance is 73\% and 81\% in AUROC when detecting two different OOD skin disease types. \cite{linmans2023b} demonstrated the highest AUROC to be ~71\% and ~81\% for two evaluations of semantic shift detection about stained histology slides. In natural image object recognition, semantic shift detection is caused by the existence of novel class instances that typically appear in a local area. However, this is not the case in some medical scenes, where the disease or lesion is typically reflected by scattered anomalies in the images. Besides, it is even difficult for a human to distinguish between semantic shiftsamples and in-distribution in some medical image modalities (e.g., endoscopic images), as their difference often lies subtle.\\

Another challenge lies in multi-label settings. Although a range of literature has explored OOD detection in supervised medical image classification, they simply assume the underlying base task to be Single Label Classification (SLC). However, it is common for multiple diseases to be present in a single medical image, which corresponds to a multi-label setting. In reality, it is impractical to train a recognition model for each disease given the huge training and annotation costs. Instead, an alternative is to train a multi-label classification (MLC) model \cite{lanchantin2021}\cite{liu2021}\cite{you2020} capable of recognizing several pre-defined diseases in a single inference run, which was recently utilized for computer-aided diagnostics \cite{zhang2023}\cite{lin2021}\cite{chen2020}. Generally, an MLC model is a neural network followed by multiple binary classification heads, each of which is responsible for judging the existence of a pre-defined class. However, most of these methods can not correctly handle out-of-distribution input during inference. In multi-label settings, a semantic shiftinstance can be either an image with only novel class instances or one with a mix of pre-defined class instances and novel class instances. For convenience, we term the first case simple OOD and the second hybrid OOD. Compared with simple OOD, it is rather difficult to distinguish a hybrid one from in-distribution samples due to the partial overlaps in their semantics. Although semantic shift detection in general image MLC has drawn attention in the research community \cite{basart2022}\cite{wang2021}\cite{wang2022}\cite{zhang2023}, they only evaluated their methods against the simple OOD. Besides, \cite{zhu2018}\cite{zhang2020} focus on the incremental learning of streaming data, where the data with new labels should be detected in each inference run. However, to our best knowledge only\cite{shi2021}\cite{wollek2023} and \cite{zhang2023} explored unseen disease detection in multi-label settings,  suggesting there is a research gap in this area.

\bibliographystyle{IEEEtran}
\bibliography{non_med_ood,med_ood}

\end{document}